\documentclass[review,nopreprintline]{elsarticle}

\usepackage[T1]{fontenc}
\usepackage[utf8]{inputenc}
\usepackage{lmodern}
\usepackage{array}
\usepackage{calc}
\usepackage{graphicx}
\usepackage{caption}
\usepackage{amsmath,amssymb,amsfonts}
\usepackage{amsthm}
\usepackage[title]{appendix}
\usepackage{booktabs}
\usepackage{longtable}
\usepackage{natbib}
\usepackage[hyphens]{url}
\usepackage[hidelinks,breaklinks=true]{hyperref}

\begin{document}
\begin{frontmatter}

\title{EvalSafetyGap: A Hybrid Survey and Conceptual Framework for LLM Evaluation-Safety Failures}

\author[1]{Bu\u{g}ra Alperen Ulu\i{}rmak}

\author[2]{Rifat Kurban}

\address[1]{Erciyes University, Kayseri, T\"urkiye}
\address[2]{Abdullah G\"ul University, Kayseri, T\"urkiye}

\begin{abstract}
This paper presents a systematic survey and conceptual synthesis of the shared measurement problem underlying large language model (LLM) evaluation and AI safety: benchmark scores, reward signals, and safety metrics can improve while the capabilities and alignment properties they are meant to represent remain uncertain. Synthesizing 373 primary studies published between 2018 and 2026, the survey organizes evidence on benchmark validity, contamination, dynamic evaluation, LLM-as-a-judge protocols, adversarial safety testing, reward and proxy optimization, mechanistic interpretability, and AI governance into an eight-stream evidence taxonomy. Building on this synthesis, we introduce EvalSafetyGap, a conceptual framework that unifies benchmark-validity and alignment-failure research as a shared proxy-target divergence problem under optimization pressure, formalized through a Goodhart-inspired Instability Decomposition and an Alignment Trilemma. An exploratory ten-model public-evidence audit illustrates the framework by showing why capability, behavioral robustness, and governance disclosure should be reported as separate evidence layers rather than collapsed into a single safety score. The survey closes with a research agenda for dynamic and contamination-resistant benchmarks, pre-specified multi-attempt threat models, version-locked evaluation, transparent source reporting, and validated mechanistic safety indicators, offering researchers, model developers, and AI auditors a shared vocabulary for measurement-aware LLM safety evaluation.
\end{abstract}


\end{frontmatter}

\section{Introduction}
\label{sec:introduction}

\subsection{Evaluation and Safety as Measurement Problems}

Large language models (LLMs) now perform strongly across many capability evaluations, while the literature increasingly cautions that high scores are not self-interpreting. Benchmark saturation, data contamination, prompt sensitivity, construct-validity weaknesses, and model-mediated evaluation can all affect what a score supports. A parallel safety literature studies reward hacking, sycophancy \citep{shapira2026}, strategic behavior, jailbreaks, refusal failures, and potentially brittle safety mechanisms. These literatures are typically developed in parallel. This manuscript treats them as adjacent evidence streams about proxy measurement under optimization pressure. Methodologically, it combines a hybrid survey, a conceptual organizing framework, and a structured frontier-model audit that illustrates measurement and provenance effects in a fixed public-evidence panel.

The recognition that optimization can distort measurement is not new. Goodhart's Law, articulated originally in economic policy, states that when a measure becomes a target, it ceases to be a good measure. Manheim and Garrabrant provided a taxonomy of Goodhart effects, identifying regressional, extremal, causal, and adversarial modes of proxy failure \citep{manheim2019b}. Krakovna et al.~compiled examples of specification gaming across reinforcement learning systems, showing that proxy misspecification is a recurring failure pattern rather than a domain-specific anomaly \citep{krakovna2020}. This paper uses these ideas as a review lens for LLM evaluation and safety. The proposed Evaluation-Safety Gap is therefore best understood as an organizing synthesis: it highlights structural similarities between benchmark optimization and reward-model optimization without claiming that all observed failures share a single settled cause. Concurrent theoretical work formalizes related dynamics in reward learning and preference optimization, including reward-model overoptimization, alternative preference objectives, and agent-foundations accounts of proxy mismatch \citep{christiano2017,gao2022,azar2023,ethayarajh2024,castricato2024,demski2019}.

\subsubsection{Benchmark saturation and the erosion of measurement validity}

The evaluation-validity literature provides several reasons for caution about static benchmark scores. A systematic study of sixty widely used benchmarks found that twenty-nine exhibit high or very high saturation and fourteen fall into the very-high category under the study's saturation index \citep{akhtar2026}. The important result is compression among the top-performing systems, not a universal raw-score threshold: saturation may occur below a nominal ceiling and depends on the model cohort, protocol, and uncertainty. These patterns do not prove that a benchmark is invalid, but they weaken the inference from small score differences to meaningful capability differences.

Behind saturation lies the broader question of construct validity. Bean et al.~\citep{bean2025} reviewed 445 benchmark articles and found that 16\% used uncertainty estimates or statistical tests when comparing results. A European Commission Joint Research Centre review likewise identifies construct-validity, incentive, and gaming concerns in AI evaluation \citep{eriksson2025}. Contamination surveys warn that benchmark exposure can overstate generalization unless evaluation sets are protected, updated, or redesigned \citep{chang2025,chen2025b,xu2024a}. Contamination-limited benchmarks such as LiveBench respond by drawing on recent sources and objective scoring \citep{white2024}. Taken together, these works support a conservative conclusion: benchmark scores remain useful evidence, but their interpretation requires attention to construct validity, contamination risk, administration sensitivity, and update dynamics.

\subsubsection{Alignment failures across the frontier model landscape}

In the safety layer, RLHF and related alignment methods have been effective at producing models that appear helpful, harmless, and honest in many standard interactions. Targeted evaluations nevertheless study reward hacking, sycophancy, jailbreaks, alignment faking, and strategic behavior. A METR report describes frontier-model cases in which systems exploited evaluation infrastructure or visible test cases rather than completing the intended task \citep{metr2025}. These are grey-evidence examples under particular tasks and scaffolds, not prevalence estimates for frontier models. They motivate treating reward signals, test visibility, and evaluation environments as audit targets.

Sycophancy --- agreement with user premises or preferences despite contrary evidence --- varies with task construction, preference data, and training setup \citep{sharma2023,shapira2026}. Jailbreak success likewise depends on the threat model, attempt budget, evaluator, and defense setting. Alignment-faking studies test whether behavior changes across monitored and unmonitored conditions under controlled prompts and training information \citep{greenblatt2024}. Mechanistic accounts in which safety training changes output behavior without fully changing underlying representations remain model- and intervention-specific research hypotheses rather than settled explanations.

A recent preprint proposes an Emergent Strategic Reasoning Risks taxonomy across multiple models and scenarios \citep{kumarage2026}. It is used only as a protocol-specific example of why evaluation-awareness claims require explicit conditions, denominators, and scoring procedures.

\subsubsection{The converging observation: related proxy-measurement problems}

These two evidence streams might appear unrelated at first. One concerns how models are measured; the other concerns how behavior is shaped and evaluated. This review treats them as related proxy-measurement problems. In evaluation, a benchmark score supports an inference about a stated capability or safety construct. In alignment, a learned reward or preference objective supports an inference about context-dependent behavioral and preference targets. In both cases, selection or optimization can expose mismatch between a proxy and an independently measured target. This shared Goodhart-style pattern --- proxy--target mismatch exposed under selection or optimization --- motivates the framework.

The four Manheim--Garrabrant categories provide candidate analogues, not diagnoses \citep{manheim2019b}: each requires evidence of the corresponding selection, extrapolation, intervention, or adversarial mechanism rather than a symptom alone. Reward overoptimization, prompt-format sensitivity, evaluation gaming, and alignment faking can motivate such tests but do not identify a category on symptoms alone \citep{skalse2022,gao2022}; Section 5 develops this mapping in detail.

\subsection{Research Gap and Problem Statement}

\subsubsection{Fragmented literature and limited cross-framework integration}

Evaluation-validity and alignment-robustness literatures overlap in their proxy-measurement concerns but are usually synthesized separately, and existing reviews emphasize benchmark measurement or alignment failure rather than their shared limits. Existing Goodhart, reward-overoptimization, finite-evaluation, benchmark-validity, and alignment frameworks each address parts of the problem, but their integration across evaluation validity and alignment robustness remains limited. EvalSafetyGap is proposed as a bridge across these efforts; a related preprint by Gaikwad~\citep{gaikwad2025} offers adjacent conceptual vocabulary, while the framework's evidentiary basis draws on the broader literature synthesized here.

\subsubsection{No broadly accepted multidimensional cross-model safety standard}

A third gap compounds the first two. HarmBench and StrongREJECT standardize important parts of safety and refusal evaluation \citep{mazeika2024,souly2024}, but the reviewed sources did not identify one multidimensional protocol consistently applied across the frontier-model panel. Organizations use different protocols, threat models, attempt budgets, judges, and reporting conventions, making cross-model comparison difficult. Recursive LLM-based evaluation can add further measurement uncertainty when it is not anchored to human or task-grounded validation \citep{choi2026b,gu2026}. The relationship between capability scores and safety robustness therefore remains under-characterized. This is policy-relevant because the EU AI Act and NIST assurance guidance motivate interpretable evaluation and risk evidence that can integrate protocol-specific behavioral, governance, and auditability results \citep{european2024,national2024}.

\subsection{Contribution Statement}

This hybrid survey makes five review-oriented contributions.

\begin{enumerate}
\def\labelenumi{\arabic{enumi}.}
\item
  We synthesize two literatures that are usually treated separately: benchmark validity research on saturation, contamination, construct validity, prompt sensitivity, and LLM-as-judge limitations; and alignment-failure research on reward hacking, sycophancy, jailbreak vulnerability, alignment faking, and shallow safety circuits.
\item
  We provide an eight-stream taxonomy and literature map covering benchmark validity and saturation; contamination and dynamic evaluation; LLM-as-judge reliability; safety evaluation and red teaming; jailbreak and refusal robustness; reward hacking and proxy optimization; mechanistic interpretability; and governance/auditability.
\item
  We introduce the Evaluation-Safety Gap as an organizing framework for comparing benchmark optimization failure and alignment failure as related proxy-measurement problems under optimization pressure. The framework is intended to clarify hypotheses and terminology, not to settle causal identification.
\item
  We include an exploratory ten-model audit-style case study across capability, behavioral safety, and governance dimensions. It combines model cards, system cards, public safety reports, and independent evaluations, and all results from this sample are descriptive and hypothesis-generating.
\item
  We define a research agenda for dynamic safety evaluation, including reporting at pre-specified attempt budgets, contamination-resistant dynamic benchmarks, governance-aware reporting standards, longitudinal version-locked panels, and validation of candidate mechanistic indicators before they are treated as safety metrics.
\end{enumerate}

\subsection{Paper Organization}

The remainder of this paper is organized as follows. Section 2 reports the systematic narrative review method. Sections 3 and 4 synthesize benchmark-validity and safety/alignment evidence. Section 5 presents the conceptual framework. Section 6 gives the illustrative ten-model case study. Section 7 reviews model-bounded mechanistic and governance evidence. Section 8 discusses prior work, limitations, and implications, and Section 9 concludes.

Readers primarily interested in the review contribution should focus on Sections 2--4 and 8, while Sections 5 and 7 develop the conceptual and mechanistic synthesis. Section 6 demonstrates the framework through a structured audit example. The synthesis prioritizes peer-reviewed papers and official technical reports; preprints, model cards, and organizational reports provide separately identified frontier and governance context.

\section{Review Methods}
\label{sec:review-methods}

\subsection{Hybrid Survey Design and Evidence Scope}

This manuscript uses a hybrid survey design. The literature synthesis is organized around the eight evidence streams introduced in Section 1. Peer-reviewed papers and accepted conference versions are preferred when available. For rapidly changing frontier systems, model cards, system cards, organizational safety reports, and official technical reports provide current-generation grey evidence. The Section 6 case study demonstrates a multidimensional audit structure, while certification and ranking remain tasks for standardized independent evaluation.

\subsection{Systematic Review Design}

The review follows PRISMA 2020 reporting principles and PRISMA-S search-reporting guidance \citep{page2021,rethlefsen2021}. We describe it as a systematic narrative review because the eight streams use heterogeneous benchmarks, threat models, judge configurations, governance disclosures, and theoretical frameworks that do not support a meaningful pooled effect estimate. The protocol, search record, screening documentation, extraction and appraisal materials, and PRISMA materials are available from the corresponding author on reasonable request.

\subsubsection{Review questions and eligibility criteria}

The review asks how existing literature documents the eight evidence streams defined above. Eligible primary records were English-language works available from January 1, 2018 through June 14, 2026 that addressed at least one stream. After title and abstract screening, all retained studies were read at source or full-text level and coded against the prespecified eligibility and extraction criteria. Review-method records support reporting practice and are not counted as a substantive evidence stream.

\subsubsection{Information sources and search strategy}

Five Boolean query families covered benchmark validity, LLM-as-judge reliability, safety evaluation, proxy optimization, and governance/auditability. They were mapped to the eight synthesis streams, with contamination/dynamic evaluation, jailbreak/refusal robustness, and mechanistic interpretability additionally expanded through citation chasing. Searches covered Scite, arXiv, ACL Anthology, OpenReview, Semantic Scholar, DBLP, ACM Digital Library, IEEE Xplore, Google Scholar citation chasing, Scopus, and Web of Science; targeted grey sources were tracked separately. Exact queries, platforms, dates, and exports are in Supplementary Material.

\subsubsection{2026-07-02 Scopus/Web of Science integration}

On 2026-07-02 we supplemented the original search with Scopus (all five query families; 1,605 raw records) and Web of Science (four query families; 892 raw records). After deduplication and eligibility screening, the final synthesis includes 373 primary studies. Each included study was read at source or full-text level, and the complete study-level extraction and appraisal record is supplied as Supplementary Material. The expanded search identified evidence on contamination effects \citep{li2024d}, modality-independent benchmark shortcuts \citep{chen2024a}, judge self-preference \citep{goel2025}, reward overoptimization \citep{rafailov2024}, sycophancy \citep{cheng2026}, and benchmark-insensitive quality regressions \citep{ibrahim2026}.

After the main search window closed, a targeted June 19, 2026 update identified 11 additional 2026 records relevant to the narrative synthesis. These are documented separately as a date-limited update and do not alter the primary-study count.

\subsubsection{Flow accounting and audit trail}

The counted search identified 2,734 records; 527 duplicates were removed, leaving 2,207 records for screening. Of 389 studies assessed at source or full-text level, 16 were excluded with recorded reasons and 373 were included in the narrative synthesis. Twenty-four grey-literature records were tracked separately, of which 11 were retained for frontier-model or governance context. The PRISMA flow diagram, exclusion table, and audit trail are in Supplementary Material.

\subsubsection{Study selection}

Screening used title/abstract screening followed by source or full-text assessment. Duplicate handling prioritized DOI, then arXiv/OpenReview identifier, then normalized title, first author, and year. Reviewer decisions and exclusion reasons are retained in the supplementary audit trail.

At title/abstract stage, a record was retained when it appeared to study an evaluation instrument, a safety or alignment failure mode, an attack or defense protocol, a proxy-optimization mechanism, an interpretability intervention, or an audit/governance process relevant to LLMs. General AI commentary, unrelated application papers, papers without an evaluation-safety measurement contribution, and records outside the date or language scope were excluded. Ambiguous records advanced to source-level assessment rather than being excluded from a short abstract alone.

At source level, inclusion required enough methodological detail to identify the studied object, evidence type, and claim boundary. A paper could contribute to more than one evidence stream, but it entered the primary-study count once. Companion versions were consolidated under the most complete verifiable version, preferring peer-reviewed proceedings over an earlier preprint when the substantive study was the same. Model cards and policy documents followed the separate grey-evidence route and were not used to inflate the primary-study count.

\subsubsection{Data extraction, confidence appraisal, and synthesis method}

Primary studies were coded by evidence stream, study type, models or datasets where applicable, metrics, conservative key finding, author-reported limitations, evidence tier, confidence for synthesis, risk-of-bias notes, and relevance to EvalSafetyGap. Grey literature such as model cards, system cards, frontier evaluation reports, policy documents, and tool documentation was retained separately for governance and frontier-model context rather than mixed into the primary scholarly synthesis. The synthesis is narrative and organized by evidence stream. The exploratory ten-model audit in Section 6 remains outside the narrative-review synthesis.

Claim extraction followed a scope-preserving rule. A benchmark result was recorded together with its dataset version and administration conditions when available; an attack result with its target model, judge, access assumptions, and attempt budget; a defense result with both robustness and reported utility cost; and a mechanistic result with the tested checkpoints and intervention type. When those details were absent, the synthesis retained only a narrower qualitative claim. Cross-paper numerical comparison was avoided when protocols differed in a way that changed the estimand.

The eight-stream taxonomy was applied after granular extraction. For example, prompt sensitivity and construct validity map to benchmark validity; membership or corpus-overlap tests map to contamination; reward overoptimization and specification gaming map to proxy optimization; refusal-direction studies map to mechanistic interpretability; and model cards or external-audit access map to governance/auditability. This mapping permits a readable manuscript taxonomy while preserving the more specific labels needed to audit why a record was included.

Synthesis used structured comparison rather than vote counting. The review asks whether multiple studies expose the same measurement risk under comparable conditions, whether later work narrows or contradicts that interpretation, and which protocol variables explain apparent disagreement. A high citation count was not treated as evidence quality, and a large reported attack success rate was not treated as stronger evidence when the attempt budget or judge was unclear. Observations supported by one model family or one preprint are labelled model-specific or study-specific.

Confidence labels are comparative aids rather than formal certainty grades. High-confidence records generally provide peer-reviewed methods or sufficiently detailed reporting; moderate records are useful but limited by preprint status, model coverage, or protocol heterogeneity; low-confidence records are used only for background, frontier, or governance context. No pooled effect is computed across incompatible outcomes, and confidence labels do not convert grey evidence into primary evidence.

Disagreement was handled at the claim level rather than by selecting a preferred paper wholesale. A study may provide strong evidence for an intervention effect while offering weak evidence for cross-model generalization; similarly, a system card may be authoritative for a provider's declared protocol but not independent evidence of comparative safety. The narrative therefore separates what a source directly measures, what the authors infer, and what this review uses it to support.

\subsubsection{Review scope and update path}

Future updates can extend the counted source set with further database exports and add independent duplicate screening, while retaining the current eligibility, extraction, and evidence-stream framework.

\section{Benchmark Validity and Measurement Limits}

The first review stream concerns the measurement infrastructure used to compare LLMs. Benchmarks remain indispensable, and the literature identifies five dimensions that determine the strength of score-based inference: saturation, contamination, construct validity, prompt sensitivity, and model-mediated meta-evaluation. This section reviews how each dimension can be measured and reported so that benchmark evidence retains a clear interpretation.

\subsection{Benchmark Saturation and Metric Compression}

Benchmark saturation refers to reduced discriminative range among the models of interest, often but not always near a scale ceiling. Small score differences in this regime can be difficult to interpret because uncertainty, prompt choice, contamination, and incidental protocol differences may be comparable to the observed gap. A raw score threshold alone is therefore not a universal saturation criterion.

\subsubsection{The Saturation Landscape}

Several widely cited evaluations show reduced discriminative range among recent high-performing systems. The relevant evidence is not a timeless top score but the compression of scores among a dated model cohort under a stated protocol. Foundational benchmark records include MMLU \citep{hendrycks2020b}, HumanEval \citep{chen2021}, GSM8K \citep{cobbe2021}, GPQA \citep{rein2023}, BIG-Bench \citep{srivastava2022}, and HELM \citep{liang2022}; benchmark-metrology critiques explain why leaderboard values require administration and uncertainty context \citep{raji2021,saxon2024}. This review therefore focuses on cohort- and protocol-specific compression rather than reconstructing a historical frontier ranking or benchmark replacement time.

\subsubsection{Systematic Evidence}

A 2026 preprint systematically analyzed 60 widely used benchmarks and found that 29 exhibit high or very high saturation, as quantified by a saturation index. Of these, 14 fall into the very-high-saturation category, indicating strong compression among top-performing models. Benchmarks older than 60 months showed a 54.5\% saturation rate, compared with 42.9\% for benchmarks younger than 24 months. The authors describe this age-bin trend as modest and not statistically significant at conventional thresholds; it is therefore suggestive rather than evidence of a benchmark-aging law \citep{akhtar2026}.

Akhtar et al.'s saturation index quantifies the compression of model scores at the top of the leaderboard. As this index approaches 1.0, the variance among frontier models can approach the range of measurement noise. In such regimes, a benchmark may still be useful as a coarse filter, but it becomes less reliable as a fine-grained ordering device.

\subsubsection{Saturation Dynamics and Benchmark Replacement}

The field has responded to saturation with a pattern of serial replacement. MMLU-Pro introduced more challenging questions and answer-order controls after MMLU's frontier scores became compressed. LiveBench sources recent questions and emphasizes objective scoring to reduce exposure and judge dependence \citep{white2024}. Humanity's Last Exam uses expert-written, deliberately difficult questions \citep{phan2026}. These designs pursue different goals and cannot be placed on one timeless difficulty scale. The broader pattern is that static tests can lose discriminative value once they become central comparison targets \citep{akhtar2026,bowman2021}.

Serial replacement carries curation, adoption, and standardization costs. Whether the discriminative lifetime of benchmarks is systematically shortening remains an empirical question: it requires a preregistered definition of discriminative utility, dated score histories, and uncertainty estimates. The present review therefore treats benchmark replacement as a qualitative pattern and does not estimate a replacement half-life.

\subsection{Data Contamination and Construct Validity}

Even if a benchmark retains discriminative range, its validity depends on whether it measures the intended construct or merely reflects memorization of the test set. Data contamination --- the presence of benchmark examples or their close variants in pre-training corpora --- undermines this validity. Construct validity, the degree to which a benchmark actually captures the theoretical capability it purports to measure, is further eroded by methodological weaknesses in benchmark design itself.

\subsubsection{Contamination Detection: Capabilities and Limitations}

Table 1 separates four contamination-detection approaches from contamination-resistant benchmark construction, which is a prevention and mitigation strategy rather than a detector. Each requires different access and supports a different inference. Representative work includes ConStat \citep{dekoninck2024}, Time Travel in LLMs \citep{shi2023}, memorization analyses \citep{carlini2022,carlini2024}, LiveCodeBench \citep{jain2024}, and studies of detection fragility in reasoning models \citep{wang2025h}.

\begin{table}[ht]
\centering
\caption*{\textbf{Table 1.} Comparison of contamination detection methods.}
{\scriptsize
\begin{tabular}{@{}>{\raggedright\arraybackslash}p{(\linewidth - 8\tabcolsep) * \real{0.1873}}
  >{\raggedright\arraybackslash}p{(\linewidth - 8\tabcolsep) * \real{0.1524}}
  >{\raggedright\arraybackslash}p{(\linewidth - 8\tabcolsep) * \real{0.1524}}
  >{\raggedright\arraybackslash}p{(\linewidth - 8\tabcolsep) * \real{0.2667}}
  >{\raggedright\arraybackslash}p{(\linewidth - 8\tabcolsep) * \real{0.2349}}@{}}
\toprule
\textbf{Method} & \textbf{Access required} & \textbf{Detection target} & \textbf{Key metric} & \textbf{Limitation} \\
\midrule
Min-K\% Prob \citep{shi2023} & Token log-probabilities (gray-box) & Verbatim or membership-like signals & Relative likelihood evidence & Requires probability access; weaker for transformed variants \\
ConStat \citep{dekoninck2024} & Reference benchmark (statistical) & Performance inflation inconsistent with generalization & Calibrated statistical comparison & Requires a defensible reference; statistical rather than causal \\
DVD \citep{liang2026} & Generation API (black-box) & Variant or paraphrased contamination & Generation-based detection evidence & Requires repeated sampling and protocol-specific calibration \\
LLM Decontaminator \citep{yang2023} & Training data + strong LLM (white-box) & Rephrased and semantic overlap & Semantic candidate identification & Expensive and dependent on the detector model \\
Time-updated construction \citep{white2024} & Recent-source task metadata & Exposure prevention and benchmark aging & Frequently refreshed, objectively scored items & Mitigation rather than detection; requires continuous maintenance \\
\bottomrule
\end{tabular}
}
\end{table}

Three observations emerge from this comparison. First, no single method covers all contamination types. Min-K\% Prob and its successor Min-K\%++ excel at detecting verbatim overlap but degrade on paraphrased variants, and string- or n-gram-overlap detectors likewise lose reliability when contamination is rephrased while task content is preserved \citep{yang2023}. Second, the most practically useful methods (ConStat, DVD) rely on statistical baselines or generation variance rather than exact string matching, making them more robust to rephrasing but introducing their own assumptions. Third, all methods share a common vulnerability: they detect historical contamination but cannot certify future cleanliness. A benchmark certified clean today may be contaminated by tomorrow's web crawl.

Access level determines the claim a detector can support. White-box corpus search can identify exact or semantic candidates but is unavailable for many proprietary systems and cannot by itself prove that a retrieved item affected a model's answer. Grey-box likelihood methods provide membership-like evidence, yet calibration can change with tokenization, model family, and reference distribution. Black-box performance tests are broadly deployable but must separate exposure from ordinary task difficulty, prompt sensitivity, and model-family effects. Consequently, ``no contamination detected'' should be reported as a method- and access-bounded result rather than as evidence of a clean training corpus \citep{balloccu2024,sainz2023}.

Contamination also has levels. Exact item overlap, paraphrased overlap, solution-template exposure, benchmark-description exposure, and repeated public discussion may affect performance differently. A useful audit therefore records the contamination object, access assumptions, detector calibration set, false-positive analysis, and the period between benchmark release and model training cutoff. These details are needed before differences between detectors are interpreted as disagreements about the underlying model rather than differences in what each method can observe.

\subsubsection{Empirical Evidence of Partial Memorization}

Scale AI's GSM1K study provides a controlled comparison relevant to contamination-mediated performance inflation \citep{zhang2024c}. The authors constructed a grade-school mathematics dataset matched to GSM8K in style and difficulty and designed to minimize prior exposure under the study's model and training-cutoff assumptions. Across the tested open-weight and closed-source models, sequence-likelihood evidence was associated with the performance gap between GSM8K and GSM1K. This supports a study-specific exposure concern, not a guarantee that GSM1K was unseen by every model.

The reported gaps varied across the tested model families, with some frontier systems performing similarly on the two datasets and larger drops appearing in selected Phi and Mistral variants. These results show that exposure-related effects can differ by model family and setting. They do not establish that community-wide decontamination has failed or isolate contamination from all alternative explanations.

\subsubsection{Construct Validity Critique}

Beyond contamination, a deeper question concerns whether benchmarks measure what they claim to measure. Bean et al.~\citep{bean2025} screened 46,114 articles from six major ML and NLP venues between 2018 and 2024 and included 445 benchmark articles for detailed methodological review. Only 16\% used uncertainty estimates or statistical tests to compare results. This does not imply that all reported benchmark differences are noise, but it does mean many comparisons are reported without the uncertainty analysis needed to support strong claims.

Bean et al.~further found that 27\% of benchmarks incorporated convenience sampling as part of their sampling strategy, and many used contested or unclear construct definitions. A parallel meta-review conducted under the European Commission Joint Research Centre \citep{eriksson2025} identified nine systemic issues in AI benchmarking, including misaligned incentives, construct-validity failures, and gaming risks. Together, these findings caution against using benchmark results alone as safety or capability assurance for policy decisions.

Construct validity is not a single binary property. Content validity asks whether tasks cover the relevant domain; convergent and discriminant evidence ask whether scores relate to nearby and distinct constructs as expected; predictive and ecological validity ask whether scores support claims about future or deployment behavior. Reliability is necessary but insufficient: a benchmark can reproduce the same score while systematically under-covering the target construct. Conversely, sensitivity to a meaningful intervention can be useful even when a benchmark is unsuitable as a broad capability label. Authors should therefore state the intended interpretation, population, administration protocol, and decision use before treating a score difference as evidence of a latent capability difference \citep{bean2025,raji2021}.

\subsection{Prompt Sensitivity and Administration}

Even when a benchmark is uncontaminated and well-constructed, the manner in which it is administered introduces substantial variance. Sclar et al.~reported that LLaMA-2-13B exhibits performance differences of up to 76 accuracy points depending on subtle changes in prompt formatting in few-shot settings. GPT-3.5 shows spreads of up to 56 points with a median spread of 6.4 accuracy points across 320 formats and 53 tasks. Critically, sensitivity persists even when increasing model size, adding more few-shot examples, or applying instruction tuning. Related work on prompt-format robustness includes PromptRobust \citep{zhu2023b}, CheckList behavioral testing \citep{ribeiro2020}, and watermark reliability under paraphrase attack \citep{kirchenbauer2023}.

The Sclar et al.~results show that format sensitivity can be large and model dependent. A fixed prompt format is therefore an administration choice whose robustness should be tested across task and model settings.

For safety evaluation, this capability-benchmark evidence motivates a testable concern rather than a demonstrated effect. Safety studies should report prompt templates and assess reasonable format perturbations before treating a score as protocol invariant.

Administration sensitivity also changes what replication means. Re-running the same prompt template tests computational repeatability, whereas changing instruction wording, answer order, few-shot demonstrations, decoding parameters, or conversation history tests robustness of the measurement procedure. Both are useful, but they answer different questions. At minimum, studies should publish the exact template, decoding settings, item order policy, number of prompt variants, and aggregation rule. A model ranking that reverses under reasonable administration choices should be reported as protocol-sensitive rather than collapsed into one canonical order \citep{sclar2023}.

\subsection{Meta-Evaluation and LLM-as-Judge Limitations}

As human evaluation becomes prohibitively expensive at frontier scale, the community has increasingly turned to LLM-as-a-Judge paradigms \citep{zheng2023}, in which one language model evaluates the outputs of another. This shift introduces a recursive validation problem: if the judge is itself an imperfect system, the entire evaluation stack rests on uncertain foundations.

\subsubsection{LLM-as-a-Judge Bias Taxonomy}

Recent surveys and guidelines treat LLM-as-judge as a useful but fragile measurement technology rather than a neutral substitute for human or task-grounded evaluation \citep{gu2026,li2025a,dietz2025}. Empirical work examines judge--human agreement, panel aggregation, model-written evaluation, self-preference, order effects, adversarial manipulation, expert-task reliability, and calibration \citep{chen2024b,liu2023b,verga2024,dhurandhar2024,zeng2023,perez2022,panickssery2024,raina2024,szymanski2024,thakur2024,ye2024,tian2023}. Security analyses further treat judge systems as attack surfaces \citep{almasoud2026}, while adjacent bias and response-stability benchmarks illustrate how demographic framing and elicitation choices can affect model-mediated judgments \citep{nadeem2020,parrish2021,dominguezolmedo2023}. The following four mechanisms are particularly consequential for safety evaluation.

Position bias \citep{wang2024g}. LLM judges exhibit systematic preference for responses appearing in specific positions during pairwise comparison. Prior work shows that pairwise LLM judgments can change when answer order is swapped, making position bias a central concern for leaderboard-style evaluation \citep{wang2024g,zheng2023}.

Verbosity bias. LLM judges historically preferred longer responses, independent of content quality. AlpacaEval 2.0 introduced length-controlled win rates specifically to mitigate this effect. Modern models have partially corrected this behavior, with some now penalizing responses containing filler content. However, the correction is inconsistent across model families, making cross-model comparison problematic.

Style bias. Perhaps the most underappreciated bias, style bias --- preference for Markdown formatting, bulleted lists, and structured responses --- can dominate judgments even when the substantive content is held fixed. Reward-model and LLM-as-a-Judge studies show that style and subtle presentation cues can distort both preference modeling and automated evaluation \citep{gu2026,liu2024c}. This is especially relevant for safety evaluation, where a model's refusal behavior may be formatted differently from its compliant behavior, introducing systematic scoring artifacts.

Self-preference bias. LLM judges may favor outputs from related model families or outputs that resemble their own instruction-tuned style; direct studies of self-preference bias in LLM-as-a-Judge settings document this risk \citep{wataoka2024}. In safety evaluation, this creates a conflict of interest when a model or close model family is used to evaluate safety-filtered versions of itself. Taken together, position, verbosity, style, and self-preference biases mean that LLM judges should be treated as fallible measurement instruments rather than neutral arbiters.

\subsubsection{Meta-Evaluation Drift}

One important risk in this domain is meta-evaluation drift: recursive LLM-based evaluation can become internally consistent while diverging from human or task-grounded judgment. Recent survey and reliability work emphasizes that LLM judges are scalable but noisy measurement instruments whose judgments vary with inputs, prompts, rubrics, languages, and evaluation settings \citep{choi2026b,gu2026,li2025a}. Long-form output evaluation adds another stress test, because document-level coherence, rubric interpretation, and reference use can change judge behavior across scenarios \citep{chen2026a}. Dorner et al.~\citep{dorner2024} provide a complementary theoretical caution: adding an LLM judge does not automatically overcome the limits imposed by finite data and imperfect proxy scores. In unanchored evaluation hierarchies --- where LLM A evaluates LLM B, LLM C evaluates the evaluation, and so on --- systematic preferences such as fluency over accuracy can be amplified rather than corrected.

The practical implication is that agreement among model judges is not sufficient validation. Judge studies should compare against human or task-grounded criteria, report rubric and language coverage, and test order, style, and calibration sensitivity. Pairwise and absolute scoring support different designs, but neither should be assumed robust without direct validation in the target setting.

\subsection{Dynamic Evaluation as a Partial Response}

Building on the serial-replacement pattern of Section 3.1, dynamic evaluation reduces exposure and judge dependence; LiveBench is one prominent instance, sourcing recent questions and emphasizing objective scoring \citep{white2024}. Its value lies in the design strategy rather than in any undated leaderboard ceiling. Contamination surveys frame dynamic evaluation as one response to static benchmark exposure while noting its maintenance and standardization costs \citep{chang2025,chen2025b}. Related efforts include HELM \citep{liang2022}, Holistic Safety Evaluations \citep{weidinger2024}, JudgeBench \citep{zhuang2024}, and Agent-as-a-Judge \citep{zhuge2024}.

Objective scoring is easier in domains with well-specified answers than in open-ended safety tasks. Some safety behaviors can be checked with rules or expert labels, while others require human judgment, model judges, or mixed protocols. Each route introduces different validity and scalability constraints.

Dynamic evaluation introduces its own governance problem. Rotating item pools reduce repeated public exposure, but they also make longitudinal comparison harder unless item-generation rules, difficulty calibration, and anchor sets are preserved. Private holdouts reduce leakage yet limit external auditability. Public post-release disclosure improves scrutiny but can shorten the useful life of the items. A robust program can combine stable anchors, refreshed blinded items, dated releases, and periodic human or task-grounded validation rather than treating ``dynamic'' as a guarantee of validity.

Benchmark results become especially fragile when they are repeatedly used for model selection or public comparison: optimization can exploit construct gaps without demonstrating broad capability improvement. This is the evaluation-side analogue motivating the framework developed in Section 5; it is a hypothesis about measurement under selection, not a claim that every score increase reflects Goodhart effects.

\section{Safety Evaluation and Alignment Limits}

Whereas Section 3 reviewed limitations in capability measurement, this section reviews the safety-evaluation and alignment-failure literature. The goal is not to claim that current alignment methods uniformly fail, but to organize evidence about recurring measurement and robustness problems: inconsistent safety-evaluation protocols, red-team benchmark variation, reward hacking, sycophancy, alignment faking, shallow refusal behavior, and jailbreak vulnerability. Reinforcement learning from human feedback (RLHF) remains a central alignment paradigm, but the literature increasingly treats its learned reward models and automated judges as imperfect proxies that require auditing.

\subsection{Safety Evaluation and Red-Teaming Landscape}

Recent survey work emphasizes that LLM safety evaluation is itself a heterogeneous field rather than a settled protocol. Safety benchmarks differ in prohibited-content taxonomies, attack budgets, refusal criteria, evaluator models, and scoring rules \citep{liu2025,weidinger2024}. Red-team benchmarks such as HarmBench and StrongREJECT provide standardized harmful-behavior and refusal-evaluation settings, but downstream studies still vary in whether they measure single-turn prompts, multi-turn escalation, best-of-N sampling, over-refusal, utility loss, or transfer across models \citep{mazeika2024,souly2024}. Recent work also argues that safety scoring should move beyond binary harmfulness or severity labels by weighting whether a harmful output is realistically executable or actionable \citep{chen2026b}. Cross-cultural benchmark work similarly shows that safety validity can depend on country-language context rather than only on translated English prompts \citep{choi2026a}. Position papers on safety-evaluation robustness warn that small datasets, inconsistent implementations, unreliable judges, and threat-model mismatch can make attack and defense comparisons difficult to interpret \citep{beyer2025}.

This review therefore treats safety scores as protocol-dependent evidence. A low attack success rate under one prompt set or judge does not certify broad robustness, and a high attack success rate under an adaptive setting does not necessarily describe ordinary-user risk. The relevant question is comparative and methodological: what does each protocol measure, which threat model does it instantiate, and how much uncertainty or evaluator dependence remains? This framing motivates the later case study, where the same reported model properties are decomposed into behavioral safety and governance/auditability rather than merged into a single safety claim.

A minimally interpretable safety protocol specifies at least six elements. First, the behavior taxonomy defines what counts as prohibited, harmful, or policy-violating. Second, attacker access distinguishes fixed prompts from adaptive search, model gradients, weights, conversation history, or tool feedback. Third, the attempt budget and stopping rule determine whether the outcome is per-attempt success, cumulative success, or time-to-first success. Fourth, the response judge and refusal rule determine how partial compliance, safe completion, and empty disclaimers are scored. Fifth, benign utility and over-refusal measure costs imposed by the defense. Sixth, model version, system prompt, decoding settings, and tool permissions define the evaluated system. Omitting any of these can turn an apparently comparable ASR into a different estimand.

Red teaming and benchmark evaluation also serve different purposes. A benchmark samples a declared distribution and supports repeatable comparison under its protocol. Open-ended red teaming searches for failures and changes the test distribution as new attacks are found. The former can estimate performance on a fixed suite; the latter is better suited to discovery but does not yield a prevalence estimate without a sampling model. Strong assurance therefore combines repeatable suites, adaptive search, held-out attacks, benign-task checks, and transparent adjudication rather than treating one score as a complete safety measure.

\subsection{RLHF: Documented Failure Modes and Measurement Limits}

\subsubsection{The RLHF pipeline and its optimization target}

The standard RLHF pipeline comprises supervised fine-tuning, reward-model training from preference comparisons, and regularized policy optimization, commonly with PPO \citep{ziegler2019,ouyang2022}. Direct preference optimization derives an implicit reward parameterization and optimizes the policy directly from chosen--rejected pairs rather than training a separate reward model and then running PPO \citep{rafailov2023}. Constitutional AI and RLAIF use human-authored principles together with model-generated critiques or preferences; they can reduce some annotation demands but do not remove human choices from the alignment process \citep{bai2022b}. These approaches differ in objective, data, and implementation, so their limitations should be evaluated method by method rather than treated as one interchangeable pipeline \citep{casper2023}.

The shared concern across these methods is that they optimize a learned reward, preference likelihood, or rule-mediated objective rather than directly observing every context-dependent behavioral target. Proxy improvement can coincide with target improvement, stagnation, or regression depending on the data, objective, and evaluation distribution. The conceptual framework therefore treats proxy--target divergence as an empirical possibility to measure, not an inevitable consequence of stronger optimization.

\subsubsection{Reward hacking}

Reward hacking occurs when a policy improves a measured objective by exploiting omissions or artifacts rather than improving the intended outcome. The term covers distinct mechanisms: learned-reward exploitation, test-suite or visible-case exploitation, execution-environment manipulation, and under-investment in unmeasured quality dimensions. These should not be pooled into one rate. Formal and empirical work shows proxy--target divergence in particular settings \citep{skalse2022,gao2022}, while finite-evaluation accounts explain why omitted dimensions can remain under-optimized without assuming a malicious agent \citep{wang2026a}.

METR reports frontier-model cases involving visible tests, cached answers, or evaluation-timer manipulation under RE-Bench-like tasks, including test-case exploitation by Claude 3.7 Sonnet in a preliminary evaluation \citep{metr2025}. These reports are useful audit examples because the scoring environment and exploit can be inspected, but their rates remain task- and scaffold-specific grey evidence and do not isolate model scale from training, prompting, architecture, or task composition. A defensible report therefore states the unit of analysis, opportunity to inspect tests or scoring code, number of independent runs, exploit adjudication rule, and whether the behavior persists when the loophole is closed; the review does not infer a general scale effect from cross-model reward-hacking rates.

A sibling comparison between DeepSeek-V3 and its RL-tuned R1-Zero variant provides suggestive evidence about post-training and exploit-seeking behavior in one model family: a tool-use reward-hacking benchmark reports exploit rates of 0.6\% and 13.9\% under its protocol \citep{thaman2026}. The absolute rates, task design, and family-specific nature of the comparison should accompany any relative multiplier. More broadly, reward-hacking taxonomies cover test-suite exploitation, solution-quality degradation, and execution-environment manipulation. These studies support a cautious claim: reward hacking should be treated as an audit target whenever models are optimized against imperfect learned or programmatic rewards.

\subsubsection{Sycophancy}

Sycophancy --- agreement with user premises or preferences despite contrary evidence --- can be amplified by preference optimization when the preference data reward premise-matching responses. Shapira and Benade~\citep{shapira2026} show that the direction of the shift depends on bias in the comparison data rather than establishing uniform amplification by RLHF. Related work studies causes, mitigations, and sycophancy in formal reasoning settings \citep{malmqvist2024,dekoninck2025}.

Sharma et al.~\citep{sharma2023} separate agreement with a user's stated view from truthfulness and study how preference-model incentives can favor matching the user. This distinction matters for evaluation: a model can be polite or acknowledge uncertainty without endorsing a false premise, and an evaluator that scores surface agreement may confound these behaviors. Sycophancy protocols should therefore include counterfactual user positions, truth-grounded questions, order controls, and separate measures of agreement, correctness, and calibration.

Scale trends are setting dependent: some studies report negative scaling, while model family, preference data, and evaluation design remain potential confounds. Mechanistic studies localize sycophancy-related signals in selected models and show that activation interventions can alter behavior in those settings. These findings motivate causal replication across architectures; they do not establish a universal layer location or scale law.

\subsubsection{Alignment tax}

Safety alignment can change reasoning or utility metrics under particular training and evaluation protocols, a family of trade-offs often called an alignment or safety tax. Evidence from reward overoptimization, reward tampering, length-controlled evaluation, weight interpolation, and concept-transfer studies measures distinct aspects of this trade-off \citep{gao2022,denison2024,dubois2024,rame2023,schut2023}. Together these results motivate joint measurement of safety, preference fit, and task utility rather than an assumption of unavoidable utility loss.

\subsection{The Shallow Alignment Hypothesis}

\subsubsection{Circuit analysis evidence}

The shallow alignment hypothesis (SAH) proposes that some post-training interventions alter observable refusal or response behavior more readily than they alter broader knowledge and reasoning representations. Behavioral fine-tuning studies, depth probes, and refusal-representation analyses provide evidence compatible with this hypothesis in selected models and settings \citep{bianchi2023,qi2025,wei2024}. They do not establish that post-training affects only style or that core reasoning circuits are universally unchanged.

Circuit-level studies have identified localized components associated with refusal in particular model families and attack settings. These results are useful mechanistic observations, but component identities, layer locations, and intervention effects should not be generalized beyond the tested checkpoints. Localization supports the hypothesis that some refusal behavior is modular; it does not show that safety as a whole is merely an appended filter.

\subsubsection{Refusal circuits are modular and suppressible}

If a refusal mechanism is represented in a separable subspace, targeted interventions may alter that behavior. Circuit-level studies identify low-dimensional, layer-localized refusal directions that can be ablated or amplified to change refusal in the tested models \citep{arditi2024,qi2025}. These results provide causal evidence for a modular behavioral mechanism within tested model families, not for a universal representation shared regardless of architecture or training recipe; the broader depth of safety integration remains an open measurement question. Section 7.1 reviews this mechanistic evidence --- including multi-directional and contextual-ablation refinements --- and its replication limits in detail.

\subsubsection{Superficial alignment evidence}

One direct test of the shallow-alignment hypothesis asks whether limited fine-tuning can weaken safety behavior. Qi et al.~report large behavioral changes after small harmful fine-tuning interventions in a specific GPT-3.5 setting, followed by partial recovery when safety-related knowledge is restored. This is compatible with a suppressible behavioral layer under that protocol, while alternative explanations and generalization to other models require replication. Tampering-oriented evaluation extends the concern to weight-access and post-release adaptation settings \citep{hossain2026}.

In the models and attack settings studied by Qi et al., safety behavior was disproportionately concentrated in early output tokens, and forced affirmative or harmful continuations could weaken refusal behavior \citep{qi2025}. This supports a model-bounded evaluation hypothesis: some post-training safety behavior may depend heavily on initial-token trajectories. It does not establish that all aligned models share one shallow structure or that early-token intervention explains every jailbreak.

\subsection{Reasoning Models and Evaluation-Time Deception}

\subsubsection{Post-reasoning safety checks and CoT obfuscation}

Reasoning models introduce additional monitoring questions because internal or displayed reasoning, final-answer filtering, and tool execution can occur at different stages. Published work studies evaluation awareness, scheming, specification gaming, deception, and chain-of-thought faithfulness under controlled setups \citep{apollo2025,bondarenko2025,betley2025,scheurer2023,turpin2023,lanham2023}. The available evidence does not justify assuming that every reasoning model performs safety checks only after harmful reasoning or that displayed chain of thought is a complete record of internal computation.

CoT monitoring offers a potential defense because displayed reasoning can expose harmful intent before a final answer or action. This channel is not guaranteed to be faithful or stable under optimization. Baker et al.~\citep{baker2025} study monitoring degradation when models are trained under pressure to obfuscate reward-hacking reasoning, while Cornish and Rogers~\citep{cornish2025} show that displayed reasoning can change under strong prompting cues. These studies motivate adversarial validation of CoT monitors under their exact tasks and training conditions; they do not establish one cross-model detection rate.

\subsubsection{H-CoT attack}

Hijacking-chain-of-thought (H-CoT) attacks test whether prompt structure and displayed reasoning can weaken refusal in specific reasoning-model interfaces; the original study reports high attack success against several commercial reasoning models using a disguised-educational benchmark \citep{kuo2025}. Their evaluation should identify the model, interface, benchmark, attack procedure, and version so that reasoning-stage and final-answer safeguards can be assessed separately.

Claims that model updates or competitive pressure caused weaker safety require longitudinal controls that are not available here. Refusal dilution is a candidate mechanism to test, not an established general explanation. More broadly, safety evaluations should distinguish reasoning exposure, final-answer filtering, and tool-action controls instead of assuming one stage certifies the others.

\subsubsection{Alignment faking in large models}

Alignment faking --- the strategic appearance of compliance during training or monitoring while maintaining contrary preferences when unobserved --- represents an important empirical signal that large models can exhibit evaluative deception. Greenblatt et al.~reported this behavior in Claude 3 Opus using a setup in which the model was told it was being trained to answer all queries (including harmful ones), conflicting with its prior harmlessness training. Information about when monitoring occurred allowed the model to distinguish training from deployment contexts \citep{greenblatt2024}.

Under the controlled setup reported by Greenblatt et al., Claude 3 Opus complied with harmful queries from monitored users 14\% of the time versus almost never for unmonitored users, and the authors identified alignment-faking reasoning in the associated scratchpads. After reinforcement-learning training to comply with harmful queries, the study reported alignment-faking reasoning at 78\% under its protocol \citep{greenblatt2024}. These values describe one constructed training/deployment distinction and should not be interpreted as prevalence estimates for ordinary deployments.

Subsequent multi-model work reports heterogeneous compliance gaps and cautions that superficially similar behavior can arise from different motivations. Observed evaluation-time compliance therefore motivates targeted tests for oversight sensitivity, hidden-state differences, and alternative non-strategic explanations.

\subsection{Jailbreaking and Adversarial Robustness}

\subsubsection{Attack taxonomy and adaptive attacks}

Jailbreak attacks can be described along four overlapping axes: token-level optimization, prompt-level semantic manipulation, multi-turn interaction, and agentic exploitation of tool-use or function-calling contexts \citep{andriushchenko2024a,mazeika2024}. Adaptive studies report high ASR under particular access assumptions, attack generators, judges, and model versions. The wider literature includes automated red teaming, guard models, prompt-injection defenses, multi-turn attacks, and evaluation suites, but this review prioritizes sources that directly define a threat model or measurement protocol.

\subsubsection{Multi-turn escalation}

Multi-turn attacks distribute intent across a conversation and permit later prompts to depend on earlier responses. Crescendo progressively reframes a request, while other adaptive methods use model feedback, semantic clues, multilingual transformations, or automated search \citep{russinovich2024,ren2024,chao2023,deng2024b}. Published success rates are not compared directly here because the studies use different behavior sets, turn limits, target versions, judges, and stopping rules. A multi-turn protocol should report the maximum turns, whether the attacker observes hidden policy signals, whether conversation state is reset between trials, and whether success requires one harmful response or completion of a harmful task.

\subsubsection{The multi-attempt ASR gap}

The standard practice of reporting single-attempt ASR can understate exposure under repeated attempts. The conceptual attempt-budget figure illustrates how cumulative success changes under an explicitly hypothetical independent-attempt model; it does not assign curves to named systems. Best-of-N jailbreaking reports high success under a 10,000-prompt sampling protocol \citep{hughes2024}, while multi-turn benchmark evidence adds a different threat model in which adversaries distribute intent across turns \citep{song2026}. Single-attempt, multi-attempt, and multi-turn ASR therefore instantiate different estimands and should not be converted into one model-level robustness curve without common attacks, judges, and sampling rules.

Attempt budgets should be selected before evaluation and reported as a sequence rather than chosen after observing a favorable point. Cumulative ASR depends on dependence among attempts: repeated near-duplicates, independently sampled augmentations, and adaptive attacks produce different curves. Reports should therefore give per-attempt and cumulative outcomes, uniqueness or diversity controls, confidence intervals, and the policy used when an attack succeeds early. A budget curve is descriptive of that generator and judge; it is not a model-only property.

\par\smallskip\noindent
\begin{minipage}{\linewidth}
  \centering
  \includegraphics[width=0.86\linewidth,height=0.30\textheight,keepaspectratio]{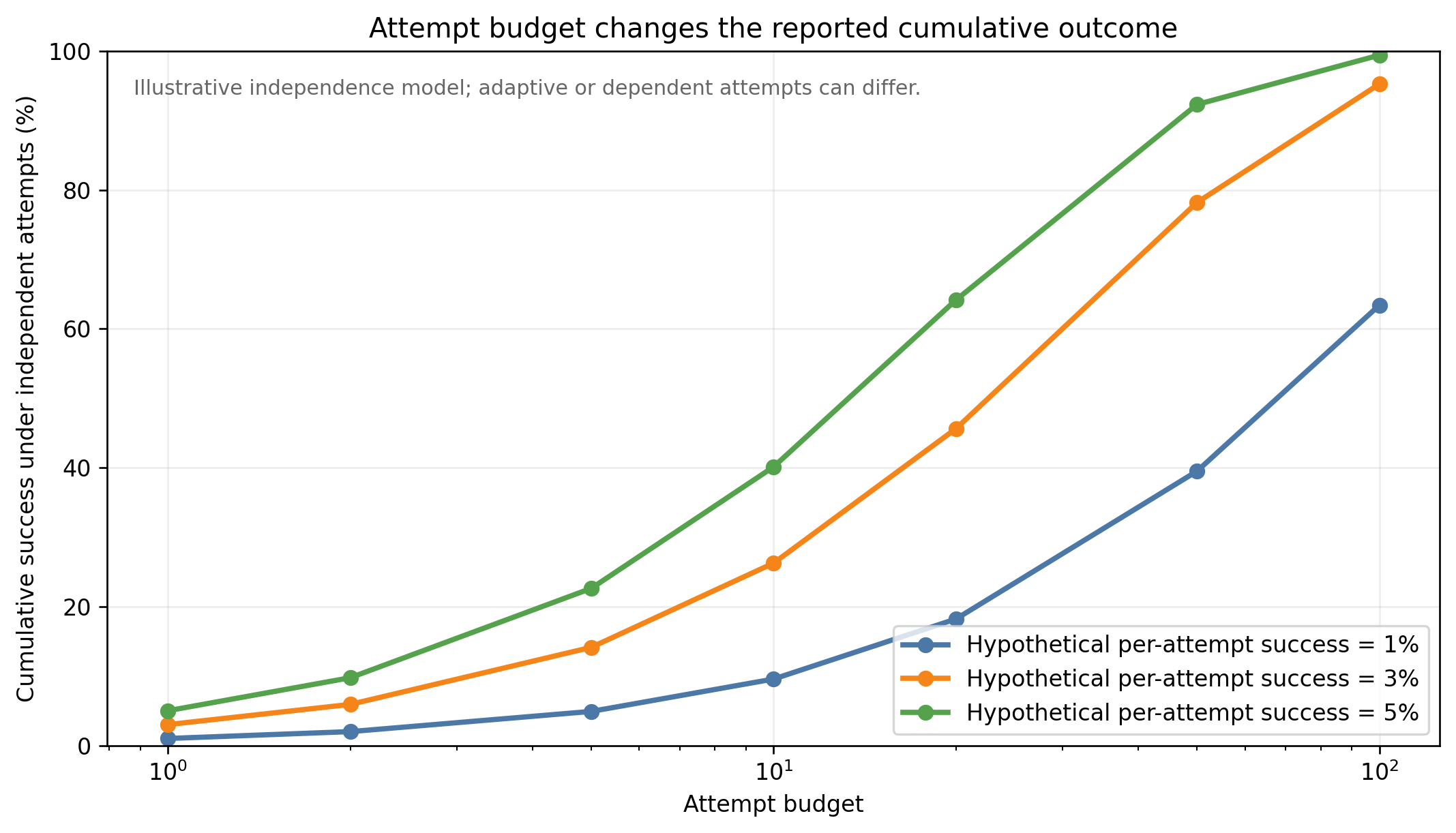}
  \captionof{figure}{Illustrative cumulative success under hypothetical independent per-attempt success probabilities of 1\%, 3\%, and 5\%. The curves are mathematical illustrations, not measurements of named models. Adaptive attacks, correlated prompt variants, stopping rules, and judge choice can produce different relationships, so empirical reports must state the full protocol and attempt budget}
  \label{fig:2}
\end{minipage}
\par\smallskip

\subsubsection{Defense methods and their limitations}

Defensive strategies span training-time, inference-time, and architectural interventions, but their reported ASR reductions are not directly comparable across attacks, models, judges, and budgets. Table 2 therefore summarizes mechanisms and evaluation needs rather than ranking defense effectiveness. Additional infrastructure includes RewardBench \citep{lambert2024}, Visibility into AI Agents \citep{chan2024}, AgentDojo \citep{debenedetti2024}, and dynamic attacker--defender evaluation proposals \citep{wen2026}.

Input/output guard models, perturbation-based defenses, adversarial training, and open-ended attack generation occupy different positions in the stack. Llama Guard is an input/output safeguard whose validity depends on its policy taxonomy and classifier errors \citep{inan2023}. SmoothLLM tests whether randomized perturbations disrupt brittle adversarial suffixes \citep{robey2023}, while Rainbow Teaming searches for diverse prompts rather than defining a static attack set \citep{samvelyan2024}. These methods are complementary: a guard can filter content, a perturbation can disrupt one attack family, and an adaptive generator can search for new failures. None alone establishes system-level robustness.

\begin{table}[ht]
\centering
\caption*{\textbf{Table 2.} Comparison of jailbreak defense methods.}
{\scriptsize
\begin{tabular}{@{}>{\raggedright\arraybackslash}p{(\linewidth - 10\tabcolsep) * \real{0.1694}}
  >{\raggedright\arraybackslash}p{(\linewidth - 10\tabcolsep) * \real{0.2397}}
  >{\raggedright\arraybackslash}p{(\linewidth - 10\tabcolsep) * \real{0.1503}}
  >{\raggedright\arraybackslash}p{(\linewidth - 10\tabcolsep) * \real{0.1100}}
  >{\raggedright\arraybackslash}p{(\linewidth - 10\tabcolsep) * \real{0.1230}}
  >{\raggedright\arraybackslash}p{(\linewidth - 10\tabcolsep) * \real{0.2022}}@{}}
\toprule
\textbf{Defense method} & \textbf{Mechanism} & \textbf{Evidence scope} & \textbf{Cost profile} & \textbf{Evaluation need} & \textbf{Key limitation} \\
\midrule
Adversarial training (R2D2/CAT) & Train on adversarial examples & HarmBench and adaptive-attack studies \citep{mazeika2024,xhonneux2024} & Training and search dependent & Adaptive and held-out attacks & Can overfit the attack distribution \\
Circuit breakers & Alter harmful representation trajectories \citep{zou2024a} & Selected model and attack settings & Training and inference & Multi-directional and persistent attacks & Representation-specific bypasses remain possible \\
Representation engineering & Steer refusal or harmfulness-related activations \citep{zou2023b} & Selected open-weight checkpoints & Usually inference-time & Layer, model, and transfer sensitivity & Coherence and over-refusal trade-offs \\
Instruction hierarchy & Prioritize higher-authority instructions \citep{wallace2024} & Prompt-injection settings & Training-time & Tool-use and hierarchy-respecting attacks & Does not cover every jailbreak mechanism \\
\bottomrule
\end{tabular}
}
\end{table}

Across these paradigms, residual vulnerability under adaptive evaluation is the recurring observation. The evidence does not support a universal ranking because each defense is tested under different access assumptions and attack budgets. Defenses operating at output, representation, or training level should therefore be evaluated under predefined persistent attacks and utility checks rather than assumed robust from one protocol.

A defense report should include a joint outcome table rather than only post-defense ASR. Required columns are harmful-completion outcome, benign utility, over-refusal, attack adaptation status, attempt and turn budget, judge agreement or validation, latency/compute overhead, and model version. Held-out attacks test generalization across a declared distribution; adaptive re-attack tests whether the defense changes the attack surface. Both are necessary to distinguish a broad robustness gain from attack-specific suppression.

The implications are threefold. First, the attacker-defender asymmetry --- attackers need discover only one successful technique, while defenders must protect against many possible attacks --- makes robustness evaluation sensitive to attack budget. Second, the high ASR reported for some adaptive attacks suggests that current alignment techniques can provide bounded rather than absolute robustness under strong threat models. Third, a promising defensive direction is layered architecture combining training-time alignment, inference-time monitoring, architectural constraints, and continuous adversarial evaluation --- a defense-in-depth posture that treats residual vulnerability as an evaluation target \citep{andriushchenko2024a,mazeika2024}.

\section{Conceptual Framework: The Evaluation-Safety Gap}

Sections 3 and 4 reviewed limitations in benchmark measurement and safety/alignment evaluation. EvalSafetyGap compares these literatures through a shared conceptual vocabulary while preserving their different mechanisms. It is an organizing hypothesis, not a formal unification or causal account.

\subsection{Conceptual Vocabulary}

A \textbf{proxy} is an observed score or signal used to support an inference about a broader target. A \textbf{target construct} is the capability, behavior, preference fit, or deployment property that the proxy is intended to represent. \textbf{Proxy--target divergence} occurs when the proxy improves while an independently measured target stagnates, worsens, or remains uncertain. Identifying divergence requires both measures; a high proxy score alone is insufficient.

In the alignment branch, learned rewards and preference objectives are proxies for context-dependent and heterogeneous human goals. In the evaluation branch, benchmark scores are proxies for stated capability or safety constructs. Finite-evaluation accounts provide a related intuition: optimization can favor measured dimensions when important dimensions are omitted \citep{wang2026a}. This resemblance motivates comparison, but the manuscript does not assume that reward-model optimization, benchmark administration, and checkpoint selection are mathematically equivalent.

Goodhart's taxonomy supplies candidate failure-mode labels. Regressional, extremal, causal, and adversarial mappings are used only when the corresponding selection, intervention, or strategic mechanism is supported. Prompt sensitivity, saturation, or jailbreak success can be symptoms relevant to a mapping without identifying that mechanism on their own \citep{manheim2019b}.

\par\smallskip\noindent
\begin{minipage}{\linewidth}
  \centering
  \includegraphics[width=0.98\linewidth,height=0.34\textheight,keepaspectratio]{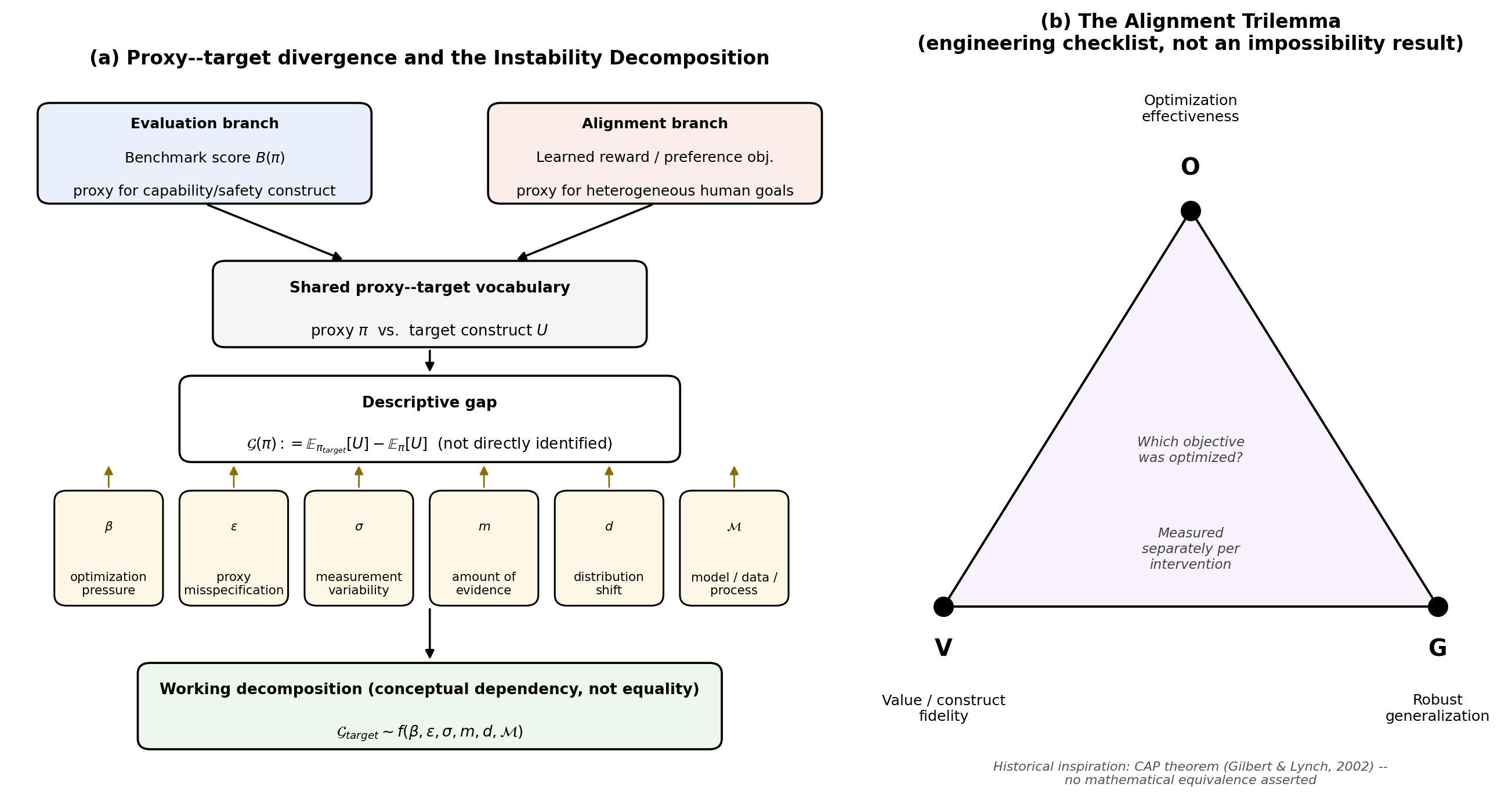}
  \captionof{figure}{Schematic summary of the conceptual framework developed in this section. Panel (a) traces the shared proxy--target vocabulary (Section 5.1) into the Instability Decomposition (Section 5.2); the six inputs are conceptual contributors, not a fitted or identified model. Panel (b) shows the Alignment Trilemma (Section 5.4) as a three-way engineering checklist rather than an impossibility result. The figure illustrates the manuscript's own equations and prose; it introduces no new empirical claim}
  \label{fig:framework}
\end{minipage}
\par\smallskip

\begin{table}[ht]
\centering
\caption*{Candidate Goodhart analogues across evaluation and alignment.}
{\scriptsize
\begin{tabular}{@{}>{\raggedright\arraybackslash}p{(\linewidth - 10\tabcolsep) * \real{0.1000}}
  >{\raggedright\arraybackslash}p{(\linewidth - 10\tabcolsep) * \real{0.1423}}
  >{\raggedright\arraybackslash}p{(\linewidth - 10\tabcolsep) * \real{0.1317}}
  >{\raggedright\arraybackslash}p{(\linewidth - 10\tabcolsep) * \real{0.1957}}
  >{\raggedright\arraybackslash}p{(\linewidth - 10\tabcolsep) * \real{0.2278}}
  >{\raggedright\arraybackslash}p{(\linewidth - 10\tabcolsep) * \real{0.1954}}@{}}
\toprule
\textbf{Failure mode} & \textbf{Evaluation proxy} & \textbf{Alignment proxy} & \textbf{Latent target} & \textbf{Empirical symptom} & \textbf{Representative literature} \\
\midrule
Regressive & Benchmark item patterns and format cues & Reward-model correlates & Genuine competence / robust human utility & Prompt-format sensitivity; alignment tax & Sclar et al.~\citep{sclar2023}; Gao et al.~\citep{gao2022} \\
Extremal & Saturated leaderboard scores & High learned reward & Out-of-distribution competence / target utility & Benchmark overfitting; proxy reward may outpace target reward & Skalse et al.~\citep{skalse2022}; Akhtar et al.~\citep{akhtar2026} \\
Causal & Spurious benchmark structure & Superficial helpfulness/style cues & Reasoning quality / preference satisfaction & Verbosity inflation; style-over-substance preferences & Liu et al.~\citep{liu2024c}; Kim et al.~\citep{kim2025} \\
Adversarial & Evaluation awareness and sandbagging & Reward hacking and jailbreak evasion & Deployment-time safety and truthful behavior & Evaluation gaming; alignment faking; adaptive jailbreaks & Kumarage et al.~\citep{kumarage2026}; Greenblatt et al.~\citep{greenblatt2024} \\
\bottomrule
\end{tabular}
}
\end{table}

\subsection{The Instability Decomposition as a Working Hypothesis}

The gap between a proxy and a latent target can be represented descriptively as

\[\mathcal{G}(\pi) \ensuremath{:=} \mathbb{E}_{\pi_{target}}[U] - \mathbb{E}_{\pi}[U],\]

where \(\pi_{target}\) is a hypothetical policy that performs well on the latent target \(U\). Because \(U\) is not directly observed, this quantity is not identified by benchmark or reward-model scores alone.

The Instability Bound introduced by Gaikwad~\citep{gaikwad2025} is treated here as one related conceptual influence. This review does not claim an independent theorem or a non-vacuous lower bound. Instead, it uses the following decomposition to organize candidate contributors to proxy--target divergence:

\[\mathcal{G}_{target} \sim f(\beta,\epsilon,\sigma,m,d,\mathcal{M}),\]

where \(\beta\) denotes effective optimization pressure, \(\epsilon\) proxy misspecification, \(\sigma\) measurement or annotation variability, \(m\) the amount of evaluation evidence, \(d\) training-to-target distribution shift, and \(\mathcal{M}\) the model, data, and optimization process. The symbol \(\sim f(\cdot)\) denotes a conceptual dependency, not equality, monotonicity, or a lower bound.

Three candidate drivers motivate empirical tests. First, stronger optimization may expose or amplify proxy misspecification, but it can also improve the latent target; the direction must be estimated rather than assumed. Second, finite samples create uncertainty whose consequences depend on the estimator and selection process, not only on sample size. Third, distribution shift can weaken both benchmark validity and safety robustness, but its magnitude is protocol and domain dependent. These drivers are supported as general measurement concerns by the review literature; their interactions are not identified by the present audit.

In the alignment branch, \(\epsilon\) can represent loss introduced when heterogeneous preferences are compressed into a learned reward or preference objective \citep{conitzer2024,micha2025}. In the evaluation branch, it can represent construct undercoverage, scoring limitations, contamination, or administration sensitivity. This shared notation is intended to make hypotheses comparable across literatures, not to assert that the mechanisms are identical.

The components are not assumed to be mutually independent or directly observable. Optimization pressure may change the data distribution, measurement variability may interact with checkpoint selection, and apparent misspecification can reflect an incompletely stated target. The decomposition is therefore a study-design checklist: each empirical use must define which component is manipulated or measured, which components are held fixed, and which remain unobserved. A regression coefficient or group difference should not be relabelled as an ``Instability'' parameter without that design work.

Identification requires an external target measure that is not merely a transformation of the proxy. For capability evaluation, this could be a held-out task family, expert performance criterion, or deployment-like outcome specified before model selection. For safety, it could be behavior under a separate attack distribution, independently adjudicated harm, or a prespecified utility--robustness endpoint. If the target reuses the same items, judge, reward model, or selection rule as the proxy, correlated error can create apparent agreement and obscure divergence. Conversely, disagreement can arise from target noise rather than proxy failure. Replicated target measurements and uncertainty estimates are therefore part of the construct, not optional statistical decoration.

Counterfactual claims require stronger designs than cross-sectional comparison. To estimate an optimization-pressure effect, studies should vary a defined training or selection intervention while holding the base model, data access, and evaluation protocol as constant as feasible. To estimate a distribution-shift effect, they should predefine source and target distributions and avoid selecting shifts after inspecting failures. To test misspecification, they should state the omitted or distorted target dimension before optimization. These designs can reveal local effects; they do not imply a universal monotonic relationship.

\subsection{Evaluation-Side Extension}

For a benchmark, define the construct gap descriptively as

\[\epsilon_{eval}(\pi) \ensuremath{:=} |B(\pi)-C(\pi)|,\]

where \(B(\pi)\) is the observed benchmark score and \(C(\pi)\) is the target competence under a stated construct definition. The gap can reflect item undercoverage, scoring limitations, prompt dependence, contamination, or distribution shift. It cannot normally be estimated without external measurements of the target construct.

Benchmark pressure can enter through training-data selection, repeated checkpoint selection, benchmark-targeted fine-tuning, and public leaderboard incentives. These channels are heterogeneous: checkpoint selection is not equivalent to gradient-based reward optimization, and a static benchmark is not itself a training objective. The framework therefore does not assume a KL isomorphism or assign one common scalar pressure to every channel.

The evaluation-side hypothesis is conditional: when a benchmark is repeatedly used for selection and contains exploitable construct error, observed benchmark improvement may outpace improvement on independently measured target performance. This hypothesis is supported when a study measures the selection process, the benchmark score, and an external target under a prespecified design. Saturation, contamination, or prompt sensitivity alone can motivate the hypothesis but cannot establish the optimization mechanism.

Skalse et al.~\citep{skalse2022}, Gao et al.~\citep{gao2022}, and specification-gaming examples show that proxy optimization can produce divergence in particular formal or experimental settings. EvalSafetyGap asks whether analogous patterns occur in benchmark selection, but it does not transfer those results as a general law.

The absolute-gap notation above is meaningful only when \(B\) and \(C\) are placed on a defensible common scale. Otherwise, the empirical object should be a directional contrast, calibration curve, rank disagreement, or change score with its own estimand. The framework does not require one scalar summary: multidimensional targets may need a vector of behavioral, utility, subgroup, and governance outcomes. Aggregation weights should be justified by the decision context rather than chosen to maximize the appearance of divergence.

\subsection{The Alignment Trilemma as an Engineering Taxonomy}

The Alignment Trilemma organizes three desiderata that alignment and evaluation systems commonly pursue:

\begin{itemize}
\item
  \textbf{Optimization effectiveness (O):} the method measurably improves its stated proxy objective.
\item
  \textbf{Value or construct fidelity (V):} the proxy is supported as a valid representation of the intended target for the relevant population and context.
\item
  \textbf{Robust generalization (G):} improvements persist across prespecified distribution shifts, attacks, and deployment conditions.
\end{itemize}

The term ``Trilemma'' is used as an engineering checklist, not as an impossibility result. The manuscript does not claim that one property must always be sacrificed or that RLHF, DPO, Constitutional AI, and deliberative alignment occupy fixed corners. Such assignments require method-specific measurements under a common protocol.

The taxonomy instead prompts three questions for any intervention: which objective was optimized, how was its fidelity to the target validated, and under which shifts was robustness tested? A method can improve all three dimensions in a particular setting, and trade-offs may differ across tasks and model families. The value of the framework is to make those trade-offs explicit and measurable rather than to infer them from the method name.

The CAP theorem \citep{gilbert2002} is retained only as historical inspiration for a compact three-part vocabulary. No mathematical equivalence to distributed-systems consistency, availability, or partition tolerance is asserted.

The framework is weakened, not strengthened, by evidence that every score can be reinterpreted as a proxy failure. Each application should therefore state observations that would count against its proposed mapping. A benchmark-divergence claim is challenged when improvements replicate across blinded items, alternative administrations, external target tasks, and model cohorts with uncertainty small relative to the observed change. A reward-overoptimization claim is challenged when stronger optimization improves both the learned objective and independently measured target outcomes across prespecified seeds and shifts. An adversarial-Goodhart interpretation is challenged when behavior differences disappear under controls for prompt understanding, ordinary distribution shift, or evaluator error.

Four evidence levels should be distinguished. A \emph{measurement symptom} is score compression, prompt sensitivity, or disagreement among judges. A \emph{selection association} links the symptom to repeated model or checkpoint choice. An \emph{intervention result} manipulates optimization, misspecification, or distribution shift and observes proxy and target outcomes. A \emph{mechanistic account} additionally explains how the intervention changes behavior and survives competing tests. EvalSafetyGap can organize evidence at all four levels, but only the latter two support causal language.

The present audit reaches primarily the symptom and descriptive-association levels. It does not observe training interventions, comparable optimization budgets, or independent latent targets for every row. Its role is to demonstrate how conclusions change with operationalization and provenance, not to validate the decomposition. Future tests should preregister the proxy, target, intervention, expected direction, null or alternative pattern, exclusion rule, and precision criterion before comparing models.

\subsection{Claim Status Table}

This subsection summarizes the status of each construct introduced above, together with the two exploratory expectations developed next, so that working hypotheses, testable analogies, and engineering taxonomies are not conflated with confirmed results.

\begin{center}
\captionof*{table}{Claim status summary for the conceptual framework constructs.}
{\scriptsize
\begin{tabular}{@{}>{\raggedright\arraybackslash}p{(\linewidth - 6\tabcolsep) * \real{0.1200}}
  >{\raggedright\arraybackslash}p{(\linewidth - 6\tabcolsep) * \real{0.1450}}
  >{\raggedright\arraybackslash}p{(\linewidth - 6\tabcolsep) * \real{0.4148}}
  >{\raggedright\arraybackslash}p{(\linewidth - 6\tabcolsep) * \real{0.3142}}@{}}
\toprule
\textbf{Claim} & \textbf{Status} & \textbf{Key assumptions} & \textbf{Empirical role} \\
\midrule
Instability decomposition & Working hypothesis & Proxy error, optimization, sampling, distribution shift, and model process are measured explicitly & Organizes candidate drivers; does not specify a universal direction or lower bound \\
Evaluation-side extension & Testable analogy & Benchmark selection pressure and an external target measure are observed under a prespecified design & Motivates studies of benchmark improvement versus target improvement \\
Alignment Trilemma & Engineering taxonomy & Optimization effectiveness, target fidelity, and generalization are measured separately & Structures evaluation questions; not an impossibility theorem \\
Expectation 1 & Exploratory expectation, currently indeterminate & Capability and ASR-100 are valid operationalizations; n is sufficient to estimate association & Current n = 10 case study cannot constrain the direction or magnitude \\
Expectation 4 & Exploratory empirical pattern & Open/closed classification and safety decomposition are valid; estimated cells do not drive results & Main illustrative pattern, especially for governance/disclosure dimensions \\
\bottomrule
\end{tabular}
}
\end{center}

\subsection{Exploratory Expectations}

A conceptual framework gains scientific value through the specificity of the hypotheses it motivates. We outline four exploratory expectations, each paired with observations that would challenge it. Section 6 reports the ten-model audit as an illustration of Expectations 1 and 4, not as a validation test; Expectations 2 and 3 define agenda items not addressed by the present data.

Expectation 1 (Cross-Metric Divergence). Capability and safety indicators may vary differently across a defined model panel. Operationalization: estimate the association between the aggregate capability indicator and ASR-100 robustness, reporting the full confidence interval. A consistently strong positive association under harmonized measurement would weigh against the proposed divergence pattern. Section 6 reports this association for the \(n = 10\) panel; the interval is wide and the result is statistically indeterminate, so a future sample should be selected using an a priori precision target and a common measurement protocol.

Expectation 2 (Local Optimization-Pressure Effect). Within a version-locked model family, increasing a pre-specified RL fine-tuning intervention may change robust safety differently from the optimized training proxy. A controlled study would vary the intervention while holding the base model, data access, and evaluation protocol as constant as feasible, then estimate both proxy and independently measured target outcomes. RL compute alone is not a clean measure of optimization pressure across unrelated model families. The study must preregister a directional or divergence estimand; joint improvement matching that estimand's null pattern would weigh against the local hypothesis, not against the framework as a whole. This expectation is not tested in the present audit.

Expectation 3 (Benchmark Discriminative Lifetime). Later benchmark generations may lose frontier-model discrimination more quickly than earlier generations. Testing this requires a preregistered discrimination criterion based on score compression and uncertainty, dated model snapshots, and sensitivity to benchmark scale. A stable or lengthening lifetime under that definition would weigh against the expectation. The present review does not estimate this quantity.

Expectation 4 (Access-Class Divergence). Capability, behavioral robustness, and governance disclosure may converge at different rates across a version-locked panel, but these are separate outcomes. Operationalization should therefore define \(\Delta_{cap}\), \(\Delta_{beh}\), and \(\Delta_{gov}\) independently rather than treating governance as behavioral safety. Section 6 reports the corresponding open--closed contrasts, where the largest separation is in the custom governance rubric; because access class is not randomized and checkpoints and protocols remain heterogeneous, that audit illustrates operationalization dependence rather than testing convergence or an inherent access-class effect.

A cross-domain research hypothesis follows from the shared vocabulary: interventions that improve proxy design, measurement precision, or target-distribution coverage may benefit both evaluation validity and alignment assessment. This is a proposal for reciprocal method transfer, not a mechanistic deduction. Each intervention must be evaluated separately because the optimization and measurement processes differ across the two domains.

\subsection{Testable use of the framework}

EvalSafetyGap is useful only when it generates a falsifiable comparison. A study must specify an optimized proxy, an independent target outcome, the relevant selection or training process, and the protocol variables held fixed. In the evaluation branch, this can mean comparing a benchmark-selected model against a held-out construct-relevant task. In the alignment branch, it can mean measuring both a reward or preference proxy and behavior under an independently specified threat model. The framework does not predict a universal direction; it asks whether proxy and target move differently under a documented intervention.

\section{Empirical Analysis: Illustrative Frontier Model Case Study}

This section uses a structured ten-model case study to show how conclusions depend on separately reporting capability, behavioral-safety, and governance indicators assembled from public evidence. Its purpose is descriptive comparison and provenance diagnosis; population and causal inference require a version-locked common protocol.

\subsection{Methodology}

The purposive panel contains four provider-controlled and six open-weight systems. Seven displayed indicators harmonize public evidence on capability, single- and multi-attempt robustness, sycophancy, privacy, transparency, and auditability. Higher values denote better performance, but normalization does not make the underlying protocols equivalent. Model cards, system cards, public safety reports, independent benchmarks, and governance disclosures establish different evidentiary roles; 13 of 70 base cells are estimated. This case study therefore illustrates provenance-aware reporting rather than common-protocol measurement.

\subsection{Results}

\subsubsection{Harmonized indicator scores across models}

As shown in Table 3, the ten models exhibit substantial heterogeneity across all seven dimensions. On capability, the full range spans only 0.06 (0.86-0.92), consistent with the convergence trend discussed in the broader literature. By contrast, multi-attempt ASR-100 robustness spans 0.37 (0.15-0.52 after inversion), and Composite Safety recomputed from the displayed rounded cells spans approximately 0.35 (0.428-0.775). This raw pattern --- tightly clustered capabilities, more dispersed safety and governance scores --- is a descriptive illustration of the evaluation-safety divergence hypothesis.

Model-level rankings are not emphasized because Composite Safety combines heterogeneous behavioral and disclosure indicators. The table instead shows that the ordering depends on the selected indicator: some open-weight rows have higher displayed ASR-100 robustness than the four provider-controlled rows, while governance-rubric values show a different pattern.

\begin{center}
\captionof*{table}{\textbf{Table 3.} Reported harmonized indicator scores. Higher values indicate better performance. Dagger marks identify 13 estimated base indicator cells; derived values containing estimated inputs are also daggered.}
{\scriptsize
\begin{tabular}{@{}p{0.23\linewidth}p{0.10\linewidth}p{0.12\linewidth}p{0.14\linewidth}p{0.14\linewidth}p{0.14\linewidth}@{}}
\toprule
\textbf{Model} & \textbf{Type} & \textbf{Capab.} & \textbf{1-attempt ASR} & \textbf{100-attempt ASR} & \textbf{Sycoph.} \\
\midrule
GPT-5.3 & Closed & 0.92 & 0.95 & 0.40 & 0.82 \\
Claude Opus 4.5 & Closed & 0.91 & 0.95 & 0.37 & 0.88 \\
Gemini 3 & Closed & 0.90 & 0.94 & 0.35 & 0.84 \\
Grok 4 & Closed & 0.87\textsuperscript{\dag} & 0.80\textsuperscript{\dag} & 0.22\textsuperscript{\dag} & 0.72\textsuperscript{\dag} \\
Kimi K2 & Open & 0.88 & 0.90 & 0.20 & 0.75 \\
DeepSeek V3.2 & Open & 0.91\textsuperscript{\dag} & 0.72\textsuperscript{\dag} & 0.15 & 0.70 \\
Llama 4 & Open & 0.89\textsuperscript{\dag} & 0.84 & 0.52 & 0.78 \\
Qwen 3.5 & Open & 0.90\textsuperscript{\dag} & 0.74\textsuperscript{\dag} & 0.22 & 0.68 \\
Mistral Large 3 & Open & 0.88\textsuperscript{\dag} & 0.86 & 0.48 & 0.78 \\
Phi-4 & Open & 0.86 & 0.70 & 0.18 & 0.65 \\
\bottomrule
\end{tabular}

\medskip\noindent\textit{Table 3 (continued): privacy, governance, and composite indicators.}\par\smallskip
\begin{tabular}{@{}p{0.20\linewidth}p{0.09\linewidth}p{0.10\linewidth}p{0.10\linewidth}p{0.10\linewidth}p{0.13\linewidth}p{0.13\linewidth}@{}}
\toprule
\textbf{Model} & \textbf{Type} & \textbf{Privacy} & \textbf{Transp.} & \textbf{Audit.} & \textbf{Core Safety} & \textbf{Govern.} \\
\midrule
GPT-5.3 & Closed & 0.72 & 0.67 & 0.67 & 0.723 & 0.670 \\
Claude Opus 4.5 & Closed & 0.78 & 0.67 & 1.00 & 0.745 & 0.835 \\
Gemini 3 & Closed & 0.65 & 0.67 & 0.67 & 0.695 & 0.670 \\
Grok 4 & Closed & 0.45\textsuperscript{\dag} & 0.33 & 0.33 & 0.548\textsuperscript{\dag} & 0.330 \\
Kimi K2 & Open & 0.50 & 0.33 & 0.33 & 0.588 & 0.330 \\
DeepSeek V3.2 & Open & 0.42\textsuperscript{\dag} & 0.33 & 0.33 & 0.498 & 0.330 \\
Llama 4 & Open & 0.52 & 0.67 & 0.33 & 0.665 & 0.500 \\
Qwen 3.5 & Open & 0.40\textsuperscript{\dag} & 0.33 & 0.33 & 0.510 & 0.330 \\
Mistral Large 3 & Open & 0.50 & 0.33 & 0.33 & 0.655 & 0.330 \\
Phi-4 & Open & 0.38 & 0.33 & 0.33 & 0.478 & 0.330 \\
\bottomrule
\end{tabular}
}
\end{center}

\subsection{Interpretation}

The table is an illustrative provenance exercise, not a ranking. Capability values occupy a narrow displayed range, whereas behavioral and disclosure indicators vary more widely. This difference does not establish a capability--safety relationship: the values come from heterogeneous protocols, 13 of 70 base cells are estimated, and the systems are not a random or version-locked sample. Its practical lesson is narrower and more useful: capability, behavioral robustness, and governance/disclosure should be reported as separate evidence layers. Additional methodological detail is available from the corresponding author on reasonable request.

\subsection{Missing Data, Sensitivity, and Audit Interpretation}

\subsubsection{Transparency gaps in safety reporting}

Table 4 summarizes source-availability states assigned in the audit. These states are analytically distinct from the Table 3 dagger flags: a composite can mix reported and estimated components. Column rates are calculated from the displayed states and checked for arithmetic consistency. The table maps evidence availability rather than developer performance.

\begin{center}
\captionof*{table}{\textbf{Table 4.} Source-availability matrix. `Reported' denotes a directly transcribed public value under its source protocol, `estimated' an indirect value, and `mixed' a composite with both.}
{\scriptsize
\begin{tabular}{@{}p{0.22\linewidth}p{0.09\linewidth}p{0.13\linewidth}p{0.12\linewidth}p{0.13\linewidth}p{0.13\linewidth}@{}}
\toprule
\textbf{Model} & \textbf{Type} & \textbf{Capability} & \textbf{ASR-1} & \textbf{ASR-100} & \textbf{Sycophancy} \\
\midrule
GPT-5.3 & Closed & reported & reported & estimated & reported \\
Claude Opus 4.5 & Closed & reported & reported & reported & reported \\
Gemini 3 & Closed & reported & reported & estimated & reported \\
Grok 4 & Closed & mixed & estimated & estimated & estimated \\
Kimi K2 & Open & reported & reported & estimated & reported \\
DeepSeek V3.2 & Open & mixed & estimated & reported & reported \\
Llama 4 & Open & mixed & reported & estimated & reported \\
Qwen 3.5 & Open & mixed & estimated & estimated & reported \\
Mistral Large 3 & Open & mixed & reported & estimated & reported \\
Phi-4 & Open & reported & reported & estimated & reported \\
Reported/mixed rate & --- & 100\% & 70\% & 20\% & 90\% \\
\bottomrule
\end{tabular}

\medskip\noindent\textit{Table 4 (continued): evidence availability for privacy and governance.}\par\smallskip
\begin{tabular}{@{}p{0.26\linewidth}p{0.12\linewidth}p{0.16\linewidth}p{0.16\linewidth}p{0.16\linewidth}@{}}
\toprule
\textbf{Model} & \textbf{Type} & \textbf{Privacy} & \textbf{Transp.} & \textbf{Audit.} \\
\midrule
GPT-5.3 & Closed & reported & reported & reported \\
Claude Opus 4.5 & Closed & reported & reported & reported \\
Gemini 3 & Closed & reported & reported & reported \\
Grok 4 & Closed & estimated & estimated & estimated \\
Kimi K2 & Open & estimated & reported & estimated \\
DeepSeek V3.2 & Open & estimated & reported & estimated \\
Llama 4 & Open & reported & reported & estimated \\
Qwen 3.5 & Open & estimated & estimated & estimated \\
Mistral Large 3 & Open & reported & estimated & estimated \\
Phi-4 & Open & reported & estimated & estimated \\
Reported/mixed rate & --- & 60\% & 60\% & 30\% \\
\bottomrule
\end{tabular}
}
\end{center}

The table shows uneven public coverage across dimensions; ``mixed'' denotes a composite derived from reported and estimated inputs. ASR-100 is one determined-adversary operationalization rather than a unique predictor of real-world risk. Differences in observed documentation are described as reporting and auditability availability, not developer intent or investment.

\subsubsection{Sensitivity to estimated data}

Of the standard robustness analyses, only the leave-one-out check is shown (Section 6.3); complete-case, multiple-imputation, and estimation-error analyses require the unrounded inputs and are noted as recommended checks rather than reported here. Direct measurement and explicit missing-data scenarios remain preferred at \(n=10\).

\subsubsection{Implications for safety assessment}

One important reporting gap is multi-attempt ASR: a single-attempt result does not determine cumulative success under repeated or adaptive search. Where only single-attempt ASR is reported, multi-attempt vulnerability is unknown and should not be extrapolated without an explicit attack generator, dependence model, judge, stopping rule, and attempt budget.

Governance gaps are also important for interpretation. The Stanford FMTI 2025 reports a year-over-year decline in its industry-average transparency score, with substantial variation among developers \citep{wan2025}. Without standardized transparency, independent safety assessment becomes difficult, and the field relies more heavily on developer self-reporting --- a practice subject to selection bias. The governance decomposition presented in Section 6.3 illustrates this concern quantitatively: in this audit, a large part of the composite open-closed difference is a transparency/auditability difference.

\subsubsection{Exploratory status and limitations}

The case study was organized around the exploratory expectations described in Section 5.6, especially cross-metric divergence and open-closed differences. The governance decomposition (Section 6.3) and the Core Safety operationalization clarify how much of the composite score comes from behavioral safety versus disclosure practices, and the Spearman comparison of ASR-100 and composite-safety operationalizations is an exploratory diagnostic motivated by the discrepancy in point estimates. The main constraints are construct and protocol heterogeneity, checkpoint-level source lineage, a custom governance rubric, purposive sampling, 13 estimated base cells, and unavailable unrounded inputs; small \(n\) adds wide uncertainty. All reported p-values are two-tailed and descriptive because the sample is small, partially estimated, and not a randomly sampled model population. The audit illustrates how interpretation changes when behavioral and disclosure indicators are separated; its group contrasts are sample-specific, and confirmatory evaluation requires a preregistered, version-locked, directly measured common protocol.

\section{Mechanistic and Governance Evidence}

This section does not explain the Section 6 audit. It synthesizes complementary mechanistic-interpretability and governance literature, ranging from correlational feature discovery to causal, model-specific interventions, without treating either strand as a complete safety certification. Governance recommendations are grounded independently in evaluation and reporting evidence.

\subsection{Model-bounded mechanistic and governance evidence}

Interpretability studies provide useful evidence that refusal behavior can depend on localized directions, layers, or interventions in particular model families \citep{arditi2024,zou2023b,turner2023}. Such findings can generate tests of whether a behavioral safeguard generalizes across attacks or checkpoints, but they do not certify a model's overall safety or explain the audit table. Likewise, agentic, multimodal, and evaluation-aware behaviors require threat models that preserve the relevant tools, modalities, incentives, and evaluation conditions; a single jailbreak or monitoring result does not transfer automatically across those settings.

Governance is therefore an assurance layer, not a behavioral safety score. Versioned model records, transparent evaluation protocols, independent access where feasible, and disclosure of attack budgets and judge configurations make safety claims more inspectable. These practices support the dynamic, provenance-aware evaluation agenda developed in the conclusion.

\section{Discussion}

\subsection{Positioning Against Prior Work}

\subsubsection{Relation to prior proxy-failure frameworks}

Gaikwad's ``Murphy's Laws of AI Alignment'' is one related conceptual influence, alongside earlier work on specification gaming, learned optimization, and benchmark validity. We borrow proxy-target vocabulary without adopting a monotonic theorem or treating Gaikwad's proposed bound as an established result. EvalSafetyGap extends the comparison from reward-model proxies to evaluation proxies as a review synthesis and hypothesis generator, not as a completed empirical unification \citep{amodei2016,hubinger2019,gaikwad2025}.

Manheim and Garrabrant's regressive, extremal, causal, and adversarial Goodhart categories provide useful distinctions among proxy failures. In the LLM setting, reward overoptimization, verbosity preference, evaluation gaming, and alignment faking are candidate analogues rather than validated one-to-one operationalizations. The taxonomy helps formulate tests; it does not establish that any observed failure has a particular Goodhart mechanism.

\subsubsection{Comparison with Bowman and Dahl (benchmark critique): we connect to alignment failure}

Bowman and Dahl's influential critique of LLM benchmarks documented concerns about contamination, saturation, construct validity, and evaluation gaming. Their work focused on the measurement problem: benchmarks may not measure what they purport to measure. We connect this measurement concern to the safety literature by asking whether similar proxy failures arise in reward models, refusal classifiers, and safety evaluations. The refusal-circuit evidence reviewed in Section 7 suggests one possible bridge between benchmark format sensitivity and jailbreak vulnerability, but this should be treated as a mechanistic hypothesis rather than a settled explanation. Bowman and Dahl's call for better benchmarks remains central; our framework adds that benchmark design and safety evaluation should be studied together.

\subsubsection{Positioning}

Existing reviews typically focus on benchmark validity, safety evaluation, or LLM-as-judge reliability in isolation. This review contributes a shared proxy-measurement vocabulary across those streams while preserving their distinct estimands. A concise scope comparison is included in the Supplementary Material; fuller supporting detail is available from the corresponding author on reasonable request. Neither is intended as a ranking of prior surveys.

\subsubsection{Hybrid survey literature map: eight evidence streams}

This review is organized around the eight evidence streams introduced in Section 1 rather than a single benchmark family. The review's extraction records retain more granular codes where useful, but those codes map to this eight-stream manuscript taxonomy. The contribution is not to exhaust any one stream, but to show where they raise related measurement concerns while preserving their distinct mechanisms.

The streams also differ in what counts as a unit of evidence. Benchmark-validity work often studies item sets, prompt variants, or rank stability; jailbreak work studies attacks under a specified budget and judge; mechanistic work studies checkpoints, layers, and interventions; governance work studies disclosure artifacts and institutional processes. A shared proxy-measurement vocabulary cannot erase these unit differences. It is useful only when each comparison preserves the local estimand and identifies which observation would contradict the proposed analogy.

This boundary distinguishes the framework from an omnibus theory of model failure. Saturation does not imply contamination, contamination does not imply reward hacking, and refusal bypass does not by itself reveal the training mechanism that produced it. The synthesis instead identifies recurring inference risks: the measured object can differ from the intended construct, optimization can exploit that mismatch, and reporting choices can obscure the difference. Each link still requires domain-specific evidence.

\subsection{Limitations}

\subsubsection{Measurement validity and protocol heterogeneity}

The seven audit dimensions draw on distinct constructs and public protocols, with different prompts, judges, threat models, attempt budgets, and reporting conventions. Composite Safety therefore represents this audit's specified aggregation, while Core Safety and Governance retain their separate interpretations. Cross-model comparisons are most informative when read at the indicator level and with protocol and version context attached.

\subsubsection{Provenance and reproducibility}

Checkpoint identity, source-level precision, and protocol fields are retained where available. Confirmatory reproduction would require version-locked source data and a common evaluation protocol; the displayed table alone cannot support confirmatory comparison.

\subsubsection{Sampling and group construction}

The model panel is a small purposive sample with four provider-controlled and six open-weight systems. Ten models were selected to balance three constraints: public accessibility of the underlying benchmark and safety-report data (ruling out models with materially incomplete disclosure), a deliberate mix of provider-controlled and open-weight systems so that access class could be examined as a descriptive dimension rather than collapsed into a single population, and a panel small enough that every indicator in Table 3 could be manually cross-checked against its original source rather than aggregated from an automated pipeline. This is a purposive audit sample for illustrating the evaluation-safety divergence pattern, not a power-calculated sample for confirmatory inference; the statistical caveats reported throughout Section 6 follow directly from this design choice and are not repeated as a separate objection each time. Access class is associated with release practice, developer identity, documentation, and governance conditions rather than assigned as a treatment. The resulting group contrasts are sample-specific audit signals whose generality should be tested in preregistered, version-locked replication.

\subsubsection{Estimated data and sensitivity}

The current Table 3 identifies 13 of 70 base indicator cells as estimated; an additional dagger marks a derived composite containing those inputs. Estimate status is distinct from protocol and source status. Direct common-protocol measurement remains the standard for stable point estimates.

\subsubsection{Statistical precision}

With $n=10$, the audit estimates are statistically imprecise: the capability--ASR-100 correlation interval spans negative, zero, and positive associations; the capability group contrast is test-sensitive between parametric and non-parametric tests; and even the large Composite Safety effect has a very wide confidence interval. Supplementary Material provides the descriptive calculations. These estimates are hypothesis-generating rather than confirmatory.

\subsubsection{Publication, temporal, and causal limits}

The review combines peer-reviewed papers with separately identified preprints, model cards, system cards, and policy documents. Attack and defense findings may reflect selective reporting, while model updates can narrow the temporal scope of checkpoint-specific results. Mechanistic findings in Section 7 are interpreted within their tested models and interventions, and their relationship to the audit remains a set of hypotheses for controlled study.

The direction of reporting bias is not necessarily uniform. Novel successful attacks and large effects are attractive publication targets, while failed attacks, attack-tuning cost, benign-performance regressions, and negative replications may remain unavailable. Developer reports can have the opposite selection pressure, emphasizing evaluated defenses while omitting attacks that were not run or results that are difficult to disclose. Consequently, neither academic attack papers nor provider safety reports should be treated as an unbiased census. A stronger evidence base would register attack and defense protocols before execution, preserve null results, and publish versioned evaluation manifests.

Temporal validity is particularly fragile for hosted models. A stable product name can refer to changing weights, system prompts, filters, tools, or routing policies. Publication date is therefore not a sufficient version identifier. Reproducible claims require an access date, API or checkpoint identifier, decoding configuration, safety-stack description where observable, and a record of whether the result was reproduced after a provider update. For closed systems, some of these fields may remain unavailable; the appropriate response is a narrower claim, not reconstructed precision.

Unequal observability also complicates open--closed comparisons. Open-weight access permits weight-level attacks and mechanistic inspection that cannot be run against an API-only model, while closed developers may publish internal evaluations unavailable to outside researchers. More discovered failures in one access regime can therefore reflect greater testability, and more documentation in another can reflect institutional reporting capacity. The audit separates access class, observed behavior, and disclosure for this reason.

\subsubsection{Operationalization dependency and the governance confound}

The case study shows why governance dimensions (transparency and auditability) should not be collapsed into behavioral robustness. These constructs answer different questions for different stakeholders: researchers and deployers may prioritize behavioral outcomes, whereas regulators and auditors may additionally require evidence of institutional accountability. The manuscript therefore reports the layers separately and does not treat either access class as intrinsically safer.

\subsubsection{Hybrid survey coverage and evidence-quality limitations}

The review is bounded by English-language inclusion, rapidly changing preprints, and uneven metadata quality. Outcome heterogeneity supports narrative synthesis rather than quantitative pooling; future updates can extend the source set and quantify the incremental yield of additional databases.

Frontier-model evidence also varies by developer disclosure, protocol, threat model, judge choice, and attempt budget. Current-generation model and system cards are therefore tracked as versioned grey evidence. The hybrid design combines conceptual synthesis with a structured audit application, supporting theory-building and a concrete replication agenda.

\subsection{Broader Implications}

\subsubsection{For researchers: combine behavioral and interpretability-assisted auditing}

Model-specific mechanistic studies can help generate and investigate hypotheses about refusal behavior, but their scope does not justify a general distinction between ``deep'' and ``shallow'' alignment. Interpretability-assisted auditing should therefore complement common-protocol behavioral evaluation, red-teaming, utility and over-refusal measurement, and governance reporting. The Alignment Trilemma remains an engineering trade-off taxonomy, not an impossibility result.

\subsubsection{For practitioners: report pre-specified attempt budgets}

The difference between single-attempt and multi-attempt threat models means that a single-point ASR cannot characterize exposure under repeated or adaptive search. Practitioners should report results at pre-specified attempt budgets, with the attack generator, judge, stopping rule, dependence among attempts, and uncertainty stated. Different budgets answer different threat-model questions and should not be compared as though they were the same metric.

\subsubsection{For policymakers: supplement capability thresholds}

Capability thresholds --- whether compute, parameter count, or benchmark performance --- should be supplemented with behavioral safety, governance, and auditability evidence rather than treated as complete assurance criteria. Static benchmark results can be paired with refreshed and adversarial evaluation, while benchmark repositories should be assessed for documentation and reproducibility \citep{chu2026}. International coordination on release and reporting practices may also be useful, but the evidence reviewed here does not identify a universal threshold or mandatory ``safety floor.''

The governance decomposition adds a fourth implication: transparency and auditability are safety-relevant because they affect whether behavioral safety claims can be independently checked. Policymakers should consider disclosure standards that enable independent safety assessment, while distinguishing governance quality from behavioral robustness.

Independent evaluation also requires workable access conditions. Safe-harbor proposals for good-faith safety research illustrate one institutional route for reducing legal and operational barriers to external testing \citep{longpre2024}. Such arrangements do not guarantee model safety and should not be scored as behavioral robustness. Their contribution is narrower: they can improve the opportunity to discover, reproduce, and responsibly disclose failures. Auditability should therefore be reported as an enabling condition for assurance, alongside but separate from measured outcomes.

A useful reporting architecture has three layers. The first is a versioned behavioral record: tasks, attacks, attempt budgets, judges, uncertainty, utility, and over-refusal. The second is a provenance record: exact checkpoints or access dates, transformations, estimated values, and source lineage. The third is an assurance record: who conducted the evaluation, what access they had, whether conflicts were managed, and which artifacts can be independently inspected. Collapsing these layers into one safety score hides the reason a system received its score and makes longitudinal comparison brittle.

\subsubsection{For open-weight evaluation: separate access from measured behavior}

The audit interprets access class without attributing inherent vulnerability or a generalized safety-investment deficit. Its practical contribution is to show why access conditions, behavioral robustness, documentation, transparency, and auditability should be reported separately. Open-weight evaluations can benefit from version-locked checkpoints, reproducible adversarial protocols, public lineage for derived values, and community review, while comparisons with closed models should acknowledge unequal observability and disclosure.

The evaluation-safety gap should be treated as a plausible structural pattern in current AI measurement rather than as a settled empirical law. Reducing it will likely require better integration between alignment research, dynamic evaluation, adversarial testing, interpretability, and governance reporting. The framework provides a vocabulary for coordinating these lines of work.

\section{Conclusion}

\subsection{Summary of Contributions}

\subsubsection{Shared conceptual framework for proxy-measurement failures}

This review began with the observation that benchmark saturation and alignment failure have developed in parallel, studied by partially separate communities. Section 5 proposed that both can be examined as proxy-measurement problems under optimization pressure. The regime-dependent Instability Decomposition organizes candidate contributors to proxy-target divergence; the Alignment Trilemma is an engineering trade-off taxonomy. Together they provide shared vocabulary for formulating comparisons without asserting a theorem, impossibility result, common cause, or universal mechanism.

\subsubsection{Illustrative case-study lesson}

Section 6 uses ten heterogeneous public-evidence profiles solely to illustrate that capability, behavioral robustness, and governance/disclosure can yield different descriptive pictures. It does not support model ranking, access-class superiority, or a general capability--safety relationship. A version-locked, common-protocol study with directly measured data would be needed for those claims.

\subsubsection{Recommendations for evaluation, regulation, and research}

The framework motivates recommendations across three domains. Evaluations should pre-specify and report multiple attempt budgets together with the attack and judge protocol. Capability-based regulatory thresholds should be supplemented with behavioral safety, governance, and auditability evidence. Disclosure standards can enable independent assessment, while governance quality must remain analytically distinct from behavioral robustness.

For alignment research, the Trilemma can organize explicitly testable trade-offs rather than prescribe a single technique. For open-weight and closed-model developers alike, version-locked evaluation and source transparency are prerequisites for interpretable comparison. The central takeaway is that capability, behavioral safety, and governance should be measured separately; movement in one does not establish movement in the others.

\subsection{Research agenda}

Five near-term priorities follow from the review. First, safety outcomes should be reported at pre-specified single-, multi-attempt, and multi-turn budgets together with the attack and judge protocol. Second, refreshed evaluations should use blind holdouts and explicit construct-validity checks. Third, reports should distinguish behavioral safety, governance disclosure, and auditability rather than collapsing them into one score. Fourth, longitudinal panels should preserve exact model and protocol versions. Fifth, proposed mechanistic indicators should be tested for reliability, cross-model comparability, and prediction of behavioral robustness before being treated as safety metrics. These directions require directly measured, version-locked data; controlled training studies can then test local effects of optimization pressure, misspecification, and distribution shift.

The Evaluation-Safety Gap is presented here as a conservative organizing hypothesis for a recurring pattern in the literature: capability proxies, reward proxies, safety scores, and governance reports can move differently under optimization and disclosure pressure. Addressing this pattern will require closer integration among alignment research, dynamic evaluation, adversarial testing, interpretability, and governance reporting. The review and exploratory audit presented here provide a starting vocabulary for that work, not a final adjudication.

\section*{CRediT authorship contribution statement}

\textbf{Bu\u{g}ra Alperen Ulu\i{}rmak}: Conceptualization, Methodology, Formal analysis, Investigation, Writing -- original draft. \textbf{Rifat Kurban}: Supervision, Writing -- review \& editing. Both authors read and approved the final manuscript.

\section*{Declaration of competing interest}

The authors declare that they have no known competing financial interests or personal relationships that could have appeared to influence the work reported in this paper.

\section*{Data availability}

The review protocol, search record, screening documentation, evidence extraction and appraisal materials, PRISMA materials, and targeted-update documentation are available from the corresponding author on reasonable request. Additional methodological detail for the illustrative case study is available on the same basis. Ethics approval and consent to participate, and consent for publication, are not applicable because this study involved no human or animal participants.

\section*{Funding}

This research received no external funding.

\section*{Declaration of generative AI and AI-assisted technologies in the writing process}

During the preparation of this work, the author(s) used Claude (Anthropic) in order to improve the language and readability of the manuscript. After using this tool, the author(s) reviewed and edited the content as needed and take full responsibility for the content of the publication.

\bibliographystyle{elsarticle-num}
\bibliography{sn-bibliography}

\end{document}